\documentclass{article} 

\usepackage[utf8]{inputenc} 
\usepackage[T1]{fontenc}    
\usepackage{hyperref}       
\usepackage{url}            
\usepackage{amsfonts}       
\usepackage{nicefrac}       
\usepackage{microtype}      

\usepackage{booktabs}
\usepackage{multirow}
\usepackage[table,xcdraw]{xcolor}         
\usepackage{graphicx}
\usepackage{xspace}
\usepackage{comment}
\usepackage{amsmath}
\usepackage{array}
\usepackage{bm}
\usepackage[algo2e,ruled,vlined,linesnumbered]{algorithm2e}
\usepackage{arydshln}

\usepackage{amssymb}
\usepackage{algorithm}
\usepackage{array}

\newcommand{\lzerohat}{\ensuremath{\hat{\ell}_0}\xspace}
\newcommand{\lzero}{\ensuremath{\ell_0}\xspace}
\newcommand{\lone}{\ensuremath{\ell_1}\xspace}
\newcommand{\ltwo}{\ensuremath{\ell_2}\xspace}
\newcommand{\linf}{\ensuremath{\ell_\infty}\xspace}
\newcommand{\lp}{\ensuremath{\ell_p}\xspace}

\newcommand{\PGDlzero}{PGD-\lzero}
\newcommand{\SPGDproj}{sPGD\ensuremath{_{p}}\xspace}
\newcommand{\SPGDunproj}{sPGD\ensuremath{_{u}}\xspace}

\newcommand{\smallcnnddn}{CNN-DDN\xspace} 
\newcommand{\smallcnntrades}{CNN-Trades\xspace} 
\newcommand{\standard}{\citep{Croce2021Robustbench}\xspace} 
\newcommand{\carmon}{\citep{carmon-19-unlabeled}\xspace} 
\newcommand{\augustin}{\citep{Augustin2020Robust}\xspace} 
\newcommand{\engstrom}{\citep{Engstrom2019robustness}\xspace} 
\newcommand{\gowal}{\citep{Gowal2021Robust}\xspace} 
\newcommand{\chen}{\citep{Chen2020Robust}\xspace} 
\newcommand{\xu}{\citep{Xu2023Robust}\xspace} 
\newcommand{\engstromimagenet}{\citep{Engstrom2019robustness}\xspace} %
\newcommand{\wong}{\citep{Wong2020Robust}\xspace} 
\newcommand{\salman}{\citep{Salman2020Robust}\xspace}
\newcommand{\resnetvulnerable}{\citep{He2015DeepRL}\xspace}
\newcommand{\hendrycks}{\citep{Hendrycks2021Robust}\xspace}
\newcommand{\debenedetti}{\citep{debenedetti2023Robust}\xspace}
\newcommand{\addepalli}{\citep{Addepalli22Robust}\xspace}
\newcommand{\crocelone}{\citep{Croce2021MindTheBox}\xspace}
\newcommand{\jianglone}{\citep{Jiang2023TowardsSA}\xspace}

\newcommand{\smallcnnddnshort}{M1\xspace} 
\newcommand{\smallcnntradesshort}{M2\xspace} 

\newcommand{\carmonshort}{C1\xspace} 
\newcommand{\augustinshort}{C2\xspace} 
\newcommand{\jiangloneshort}{C4\xspace} 
\newcommand{\standardshort}{C5\xspace} 
\newcommand{\gowalshort}{C6\xspace} 
\newcommand{\engstromshort}{C7\xspace} 
\newcommand{\chenshort}{C8\xspace} 
\newcommand{\xushort}{C9\xspace} 
\newcommand{\addepallishort}{C10\xspace} 
\newcommand{\croceloneshort}{C3\xspace} 
\newcommand{\zhongsatshort}{C12\xspace} 
\newcommand{\zhongtradesshort}{C11\xspace} 
\newcommand{\standardmed}{\standardshort~\citep{Croce2021Robustbench}\xspace} 
\newcommand{\carmonmed}{\carmonshort~\citep{carmon-19-unlabeled}\xspace} 
\newcommand{\augustinmed}{\augustinshort~\citep{Augustin2020Robust}\xspace} 
\newcommand{\engstrommed}{\engstromshort~\citep{Engstrom2019robustness}\xspace} 
\newcommand{\gowalmed}{\gowalshort~\citep{Gowal2021Robust}\xspace} 
\newcommand{\chenmed}{\chenshort~\citep{Chen2020Robust}\xspace} 
\newcommand{\xumed}{\xushort~\citep{Xu2023Robust}\xspace} 
\newcommand{\addepallimed}{\addepallishort~\citep{Addepalli22Robust}\xspace}
\newcommand{\zhongsatmed}{\zhongsatshort~\citep{zhong2024towards}\xspace}
\newcommand{\zhongtradesmed}{\zhongtradesshort~\citep{zhong2024towards}\xspace}

\newcommand{\resnetvulnerableshort}{I1\xspace}
\newcommand{\engstromimagenetshort}{I2\xspace} 
\newcommand{\hendrycksshort}{I3\xspace}
\newcommand{\debenedettishort}{I4\xspace}
\newcommand{\wongshort}{I5\xspace}
\newcommand{\salmanshort}{I6\xspace}
\newcommand{\moshort}{I8\xspace}
\newcommand{\pengshort}{I7\xspace}

\newcommand{\resnetvulnerablemed}{\resnetvulnerableshort~\citep{He2015DeepRL}\xspace}
\newcommand{\engstromimagenetmed}{\engstromimagenetshort~\citep{Engstrom2019robustness}\xspace} 
\newcommand{\wongmed}{\wongshort~\citep{Wong2020Robust}\xspace} 
\newcommand{\salmanmed}{\salmanshort~\citep{Salman2020Robust}\xspace}
\newcommand{\hendrycksmed}{\hendrycksshort~\citep{Hendrycks2021Robust}\xspace}
\newcommand{\debenedettimed}{\debenedettishort~\citep{debenedetti2023Robust}\xspace}
\newcommand{\momed}{\moshort~\citep{mo2022adversarial}\xspace}
\newcommand{\pengmed}{\pengshort~\citep{peng2023adversarial}\xspace}

\newcommand{\medianlzero}{\ensuremath{\mathbf{\tilde{\ell}}_0}\xspace}

\NewDocumentCommand{\rot}{O{45} O{1em} m}{\makebox[#2][l]{\rotatebox{#1}{#3}}}%

\newcommand{\sigmazero}{\ensuremath{\mathtt{\sigma}}\texttt{-zero}\xspace}

\newcommand{\vct}[1]{\boldsymbol{\mathbf{#1}}}
\newcommand\set[1]{\mathcal{#1}}

\newcommand{\binfty}{\ensuremath{\boldsymbol{\textcolor{gray}{\infty}}}}
\newcommand{\minfty}{\textbf{\textcolor{gray}{ > 40}}}

\SetKwInput{KwInput}{Input}                
\SetKwInput{KwOutput}{Output}
\SetKwComment{tcp}{$\triangleright$\ }{}

\SetCommentSty{mycommfont}

\newcommand{\myparagraph}[1]{\noindent \textbf{#1}}

\newcommand{\rebuttal}[1]{#1}

\newcommand{\corresponding}[1]{\let\thefootnote\relax\footnotetext{$^\dagger$#1}}

    \makeatletter
\newcommand{\linebreakand}{%
  \end{@IEEEauthorhalign}
  \hfill\mbox{}\par
  \mbox{}\hfill\begin{@IEEEauthorhalign}
}
\makeatother

\newcommand{\tcheck}{\checkmark}

\newcommand{\cc}{\cellcolor{gray!10}}
\newcommand{\rc}{\rowcolor{blue!10}}

\newcommand{\sigmazeroES}{\ensuremath{\mathtt{\sigma}}\texttt{-zero}$_k$\xspace}

\newcommand{\email}[1]{\footnotesize{\texttt{\textcolor{magenta}{#1}}}}

\usepackage{preprint,times}

\usepackage{amsmath,amsfonts,bm}

\def\eqref#1{equation~\ref{#1}}

\def\1{\bm{1}}

\DeclareMathAlphabet{\mathsfit}{\encodingdefault}{\sfdefault}{m}{sl}
\SetMathAlphabet{\mathsfit}{bold}{\encodingdefault}{\sfdefault}{bx}{n}

\DeclareMathOperator{\sign}{sign}

\usepackage{hyperref}
\usepackage{url}

\usepackage[utf8]{inputenc} 
\usepackage[T1]{fontenc}    
\usepackage{booktabs}       
\usepackage{amsfonts}       
\usepackage{nicefrac}       
\usepackage{microtype}      
\usepackage{cleveref}

\Crefname{equation}{Eq.}{Eqs.}

 \Crefrangeformat{equation}{Eqs.~(#3#1#4)-(#5#2#6)}

\title{$\sigma$-zero: Gradient-based Optimization of \\$\ell_0$-norm Adversarial Examples}

\author{Antonio Emanuele Cinà\textsuperscript{1} 
\qquad ~~Francesco Villani\textsuperscript{1} \qquad ~~Maura Pintor\textsuperscript{2} \qquad ~~Lea Schönherr\textsuperscript{3}\\ \textbf{Battista Biggio}\textsuperscript{2} \qquad ~~\textbf{Marcello Pelillo\textsuperscript{4}} \\
\textsuperscript{1}Department of Computer Science, Bioengineering, Robotics and Systems, University of Genoa, Italy\\ 
\textsuperscript{2}Department of Electrical and Electronic Engineering, University of Cagliari, Italy\\
\textsuperscript{3}CISPA Helmholtz Center for Information Security, Germany\\
\textsuperscript{4}Department of Environmental Sciences, Informatics and Statistics, Ca' Foscari University of Venice, Italy\\
\email{antonio.cina@unige.it} \quad \email{francesco.villani@edu.unige.it} \quad\email{maura.pintor@unica.it} \\ \email{schoenherr@cispa.de} \quad \email{battista.biggio@unica.it} \quad \email{pelillo@unive.it}
}

\newcommand\blfootnote[1]{%
  \begingroup
  \renewcommand\thefootnote{}\footnote{#1}%
  \addtocounter{footnote}{-1}%
  \endgroup
}

\iclrfinalcopy 
\begin{document}

\maketitle

\begin{abstract}
Evaluating the adversarial robustness of deep networks to gradient-based attacks is challenging.
While most attacks consider $\ell_2$- and $\ell_\infty$-norm constraints to craft input perturbations, only a few investigate sparse $\ell_1$- and $\ell_0$-norm attacks.
In particular, $\ell_0$-norm attacks remain the least studied due to the inherent complexity of optimizing over a non-convex and non-differentiable constraint.
However, evaluating adversarial robustness under these attacks could reveal weaknesses otherwise left untested with more conventional $\ell_2$- and $\ell_\infty$-norm attacks.
In this work, we propose a novel $\ell_0$-norm attack, called $\sigma$\texttt{-zero}, which leverages a differentiable approximation of the $\ell_0$ norm to facilitate gradient-based optimization, and an adaptive projection operator to dynamically adjust the trade-off between loss minimization and perturbation sparsity.
Extensive evaluations using MNIST, CIFAR10, and ImageNet datasets, involving robust and non-robust models, show that $\sigma$\texttt{-zero} finds minimum $\ell_0$-norm adversarial examples without requiring any time-consuming hyperparameter tuning, and that it outperforms all competing sparse attacks in terms of success rate, perturbation size, and efficiency.
\end{abstract}

\section{Introduction}
Early research has revealed that machine learning models are fooled by adversarial examples, i.e., slightly-perturbed inputs optimized to cause misclassifications~\citep{Biggio13Evasion,szegedy-14-intriguing}. The discovery of this phenomenon has, in turn, demanded a more careful evaluation of the robustness of such models, especially when deployed in security-sensitive and safety-critical applications.
Most of the gradient-based attacks proposed to evaluate the adversarial robustness of Deep Neural Networks (DNNs) optimize adversarial examples under different $\ell_p$-norm constraints. In particular, while convex \lone, \ltwo, and \linf norms have been widely studied~\citep{Chen2018EAD,Croce2021MindTheBox}, only a few \lzero-norm attacks have been considered to date.
The main reason is that finding minimum $\ell_0$-norm solutions is known to be an NP-hard problem~\citep{davis1997adaptive}, and thus ad-hoc approximations must be adopted to overcome issues related to the non-convexity and non-differentiability of such (pseudo) norm.
Although this is a challenging task, attacks based on the \lzero norm have the potential to uncover issues in DNNs that may not be evident when considering other attacks~\citep{Carlini17CW,Croce2021MindTheBox}. In particular, \lzero-norm attacks, known to perturb a minimal fraction of input values, can be used to determine the most sensitive characteristics that influence the model's decision-making process, offering a different and relevant threat model to benchmark existing defenses and a different understanding of the model's inner workings. {\hypersetup{hidelinks}\blfootnote{Code is available at \textcolor{magenta}{\url{https://github.com/sigma0-advx/sigma-zero}}.}}

%
\begin{figure*}[t]
\centering\includegraphics[width=0.97\textwidth]{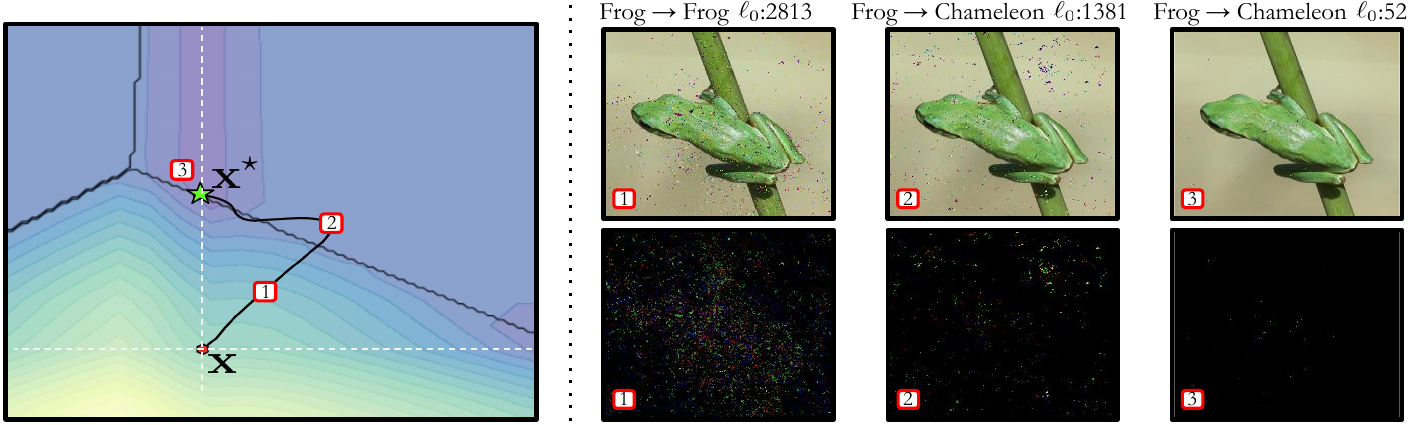}
    \caption{The leftmost plot shows the execution of \sigmazero on a two-dimensional problem. The initial point $\vct{x}$ (\textit{red dot}) is updated via gradient descent to find the adversarial example $\vct{x}^\star$ (\textit{green star}) while minimizing the number of perturbed features (i.e., the \lzero norm of the perturbation). The gray lines surrounding $\vct{x}$ demarcate regions where the \lzero norm is minimized. The rightmost plot shows the adversarial images (\textit{top row}) and the corresponding perturbations (\textit{bottom row}) found by \sigmazero during the three steps highlighted in the leftmost plot, along with their prediction and \lzero norm.}
    \label{fig:abstract_figure}
\end{figure*}

Unfortunately, current \lzero-norm attacks exhibit a largely suboptimal trade-off between their success rate and efficiency, i.e., they are either accurate but slow or fast but inaccurate. 
In particular, the accurate ones use complex projections and advanced initialization strategies (e.g., adversarial initialization) to find smaller input perturbations but suffer from time or memory limitations, hindering their scalability to larger networks or high-dimensional data~\citep{Brendel2019BB,Cesaire2021VFGA}.
Other attacks execute faster, but their returned solution is typically less accurate and largely suboptimal~\citep{Matyasko2021PDPGD,Pintor2021FMN}. 
This results in overestimating adversarial robustness and, in turn, contributes to spreading a \textit{false sense of security},  hindering the development of effective defense mechanisms~\citep{Carlini2019EvalRobustness,pintor2022indicators}. 
Developing a reliable, scalable, and compelling method to assess the robustness of DNN models against sparse perturbations with minimum \lzero norm remains thus a relevant and challenging open problem.

%
%
In this work, we propose a novel $\ell_0$-norm attack, named \sigmazero, which iteratively promotes the sparsity of the adversarial perturbation by minimizing its \lzero norm (see 
\autoref{fig:abstract_figure} and \autoref{sec:methodology}). 
To overcome the limitations of previous approaches, our attack leverages two main technical contributions:
(i) a smooth, differentiable approximation of the \lzero norm to enable the minimization of the attack loss via gradient descent; and (ii) an adaptive projection operator that dynamically increases sparsity to further reduce the perturbation size while keeping the perturbed sample in the adversarial region.

Our experiments (\autoref{sec:experiments}) provide compelling evidence of the remarkable performance of \sigmazero. We evaluate it on 3 well-known benchmark datasets (i.e., MNIST, CIFAR10, and ImageNet), using \rebuttal{22} different models from Robustbench~\citep{Croce2021Robustbench} and the corresponding official repositories. 
We compare the performance of \sigmazero against more than $10$ competing attacks, totaling almost 450 different comparisons. Our analysis shows that \sigmazero outperforms state-of-the-art attacks in terms of both attack success rate and perturbation size (lower \lzero norm), while being also significantly faster (i.e., requiring fewer queries and lower runtime). 
Our attack also provides some additional advantages: (i) it does not require any sophisticated, time-consuming hyperparameter tuning; (ii) it does not require being initialized from an adversarial input; (iii) it is less likely to fail, i.e., it consistently achieves an attack success rate of 100\% for sufficiently-large perturbation budgets, thereby enabling more reliable robustness evaluations~\citep{Carlini2019EvalRobustness}. 
We thus believe that \sigmazero will foster significant advancements in the development of better robustness evaluation tools and more robust models against sparse attacks.
We conclude the paper by discussing related work (\autoref{sec:relatedwork}), along with the main contributions and future research directions (\autoref{sec:conclusion}).
\section{\sigmazero: Minimum $\ell_0$-norm Attacks}\label{sec:methodology}
We present here \sigmazero, a gradient-based attack that finds minimum \lzero-norm adversarial examples. 

\myparagraph{Threat Model.} We assume that the attacker has complete access to the target model, including its architecture and trained parameters, and exploits its gradient for staging white-box untargeted attacks~\citep{Carlini17CW,Biggio18WildPatterns}. 
This setting is useful for worst-case evaluation of the adversarial robustness of DNNs, providing an empirical assessment of the performance degradation that may be incurred under attack. Note that this is the standard setting adopted in previous work for gradient-based adversarial robustness evaluations~\citep{Carlini17CW,brendel-19-accurate,Croce2021Robustbench,Pintor2021FMN}.

\myparagraph{Problem Formulation.} In this work, we seek untargeted minimum \lzero-norm adversarial perturbations that steer the model's decision towards misclassification~\citep{Carlini17CW}. To this end, let $\vct x \in \set X = [0,1]^d$ be a $d$-dimensional input sample, $y \in \set Y = \{1, \ldots, l\}$ its associated true label, and $f: \set X \times \Theta \mapsto \set Y$ the target model, parameterized by $\vct \theta \in \Theta$. 
While $f$ outputs the predicted label, we will also use $f_k$ to denote the continuous-valued output (logit) for class $k \in \set Y$.
The goal of our attack is to find the minimum \lzero-norm adversarial perturbation $\vct \delta^\star$ such that the corresponding adversarial example $\vct x^\star = \vct x + \vct \delta^\star$ is misclassified by $f$. This can be formalized as:
\begin{align}
    \vct\delta^\star \in \arg\min_{\vct \delta} & \quad \|\vct\delta\|_0 \, , \label{eq:attack-obejctive}\\ 
    \textrm{s.t.} & \quad f(\vct x + \vct \delta, \vct \theta) \neq y \, , \label{eq:adv-constraint}\\ 
    & \quad  \vct x + \vct \delta \in [0, 1]^d  \label{eq:box-constraints} \, ,
\end{align}
where $\|\cdot\|_0$ denotes the \lzero norm, which counts the number of non-zero components. The hard constraint in \Cref{eq:adv-constraint} ensures that the perturbation $\vct \delta$ is valid only if the target model $f$  misclassifies the perturbed sample $\vct x + \vct \delta$, while the box constraint in \Cref{eq:box-constraints} ensures that the perturbed sample lies in $[0,1]^d$.\footnote{Note that, when the source point $\vct x$ is already misclassified by $f$, the solution is simply $\vct \delta^\star= \vct 0$.}
Since the problem in \Crefrange{eq:attack-obejctive}{eq:box-constraints} can not be solved directly, we reformulate it as:
\begin{align}
    \vct\delta^\star \in \arg\min_{\vct \delta}& \quad  \set{L}(\vct x + \vct \delta, y, \vct \theta) +  \frac{1}{d}\hat{\ell}_0(\vct\delta) \label{eq:attack-obejctive-approximated}\\ 
     \textrm{s.t.} & \quad  \vct x + \vct \delta \in [0, 1]^d \, , ~\label{eq:box-constraints-2}
\end{align}
where we use a differentiable approximation $\hat{\ell}_0(\vct\delta)$ instead of $||\vct \delta ||_0$, and normalize it with respect to the number of features $d$ to ensure that its value is within the interval $[0,1]$. The loss $\set{L}$ is defined as: 
\begin{align}
\set{L}(\vct x, y, \vct \theta) = \max \left ( f_y(\vct x, \vct \theta) - \max_{k\neq y} f_k(\vct x, \vct \theta) , 0 \right)  +\mathbb{I}(f(\vct x, \vct \theta) = y) \, .
\label{eq:sigma-zero-loss-new}
\end{align}
The first term in $\set{L}$ represents the logit difference, which is positive when the sample is correctly assigned to the true class $y$, and clipped to zero when it is misclassified~\citep{Carlini17CW}. The second term merely adds $1$ to the loss if the sample is correctly classified.\footnote{While a sigmoid approximation may be adopted to overcome the non-differentiability of the $\mathbb{I}$ term at the decision boundary, we simply set its gradient to zero \textit{everywhere}, without any impact on the experimental results.}
This ensures that $\set{L}=0$ only when an adversarial example is found and $\set{L}\geq1$ otherwise. 
In practice, when minimizing the objective in \Cref{eq:attack-obejctive-approximated}, this loss term induces an \textit{alternate} optimization process between minimizing the loss function itself (to find an adversarial example) and minimizing the $\ell_0$-norm of the adversarial perturbation (when an adversarial example is found).
It is also worth remarking that, conversely to the objective function proposed by~\citet{Carlini17CW}, our objective does not require tuning any trade-off hyperparameters to balance between minimizing the loss and reducing the perturbation size, thereby avoiding a computationally expensive line search for each input sample.

\myparagraph{$\ell_0$-norm Approximation.} Besides the formalization of the attack objective, one of the main technical advantages of \sigmazero is the smooth, differentiable approximation of the $\ell_0$ norm, thereby enabling the use of gradient-based optimization.
To this end, we first note that the \lzero-norm of a vector can be rewritten as $\|\vct x \|_0 = \sum_{i=1}^d \sign(x_i)^2$, and then approximate the \emph{sign} function as $\sign(x_i) \approx {x_i}/{\sqrt{x_i^2 + \sigma}}$, where $\sigma >0$ is a smoothing hyperparameter that makes the approximation sharper as $\sigma \rightarrow 0$. This, in turn, yields the following smooth approximation of the \lzero norm:
\begin{align}
    \hat{\ell}_0(\vct x, \sigma) = \sum\limits_{i=1}^{d} \frac{x_i^2}{x_i^2+\sigma}, &  \sigma > 0, \quad \hat{\ell}_0(\vct x, \sigma) \in [0, d] \,.
    \label{eq:l0_approximation}
\end{align}

\myparagraph{Adaptive Projection $\Pi_\tau$.} The considered \lzero-norm approximation allows optimizing \Cref{eq:attack-obejctive-approximated} via gradient descent. 
However, using such a smooth approximation tends to promote solutions that are not fully sparse, i.e., with many components that are very close to zero but not exactly equal to zero, thereby yielding inflated \lzero-norm values. 
To overcome this issue, we introduce an adaptive projection operator $\Pi_\tau$ that sets to zero the components with a perturbation intensity lower than a given \textit{sparsity threshold} $\tau$ in each iteration. The sparsity threshold $\tau$ is initialized with a starting value $\tau_0$ and then dynamically adjusted for each sample during each iteration; in particular, it is increased to find sparser perturbations when the current sample is already adversarial, while it is decreased otherwise. The updates to $\tau$ are proportional to the step size and follow its annealing strategy, as detailed below. 

\myparagraph{Solution Algorithm.} 
Our attack, given as \autoref{alg:sigma-zero}, solves the problem in \Crefrange{eq:attack-obejctive-approximated}{eq:box-constraints-2} via a fast and memory-efficient gradient-based optimization. 
After initializing the adversarial perturbation $\vct{\delta}=\vct 0$ (\autoref{line:initialization}), it computes the gradient of the objective in \Cref{eq:attack-obejctive-approximated} with respect to $\vct{\delta}$ (\autoref{line:objective_gradient}). The gradient is then normalized such that its largest components (in absolute value) equal $\pm 1$ (\autoref{line:gradient_normalization}). This stabilizes the optimization by making the update independent from the gradient size, and also makes the selection of the step size independent from the input dimensionality~\citep{Rony2018DecouplingDA,Pintor2021FMN}. 
%
We then update $\vct \delta$ to minimize the objective via gradient descent while also enforcing the box constraints in \Cref{eq:box-constraints-2} through the usage of the \texttt{clip} operator (\autoref{line:final_update}).
We increase sparsity in $\vct\delta$ by zeroing all components lower than the current sparsity threshold $\tau$ (\autoref{line:tau_threshold_enforcement}), as discussed in the previous paragraph.
We then decrease the step size $\eta$ via cosine annealing (\autoref{line:cosinescheduler}), as suggested by~\citet{Rony2018DecouplingDA,Pintor2021FMN}, and adjust the sparsity threshold $\tau$ accordingly (\autoref{line:tau_threshold_increase}). In particular, if the current sample is adversarial, we increase $\tau$ by $t \cdot \eta$ to promote sparser perturbations; otherwise, we decrease $\tau$ by the same amount to promote the minimization of $\set L$. 
The above process is repeated for $N$ iterations while keeping track of the best solution found, i.e., the adversarial perturbation $\vct \delta^\star$ with the lowest \lzero norm (\autoref{line:best_delta}). 
If no adversarial example is found, the algorithm sets $\vct \delta^\star = \infty$ (\autoref{line:best_delta_infty}). It terminates by returning $\vct x^\star=\vct x + \vct \delta^\star$ (\autoref{line:end_attack}).

\begin{algorithm}[t]
  \SetAlgoLined
\DontPrintSemicolon
  \KwInput{$\vct x\in [0,1]^d$, the input sample; y, the true class label; $\vct\theta$, the target model; N,~the number of iterations;   $\eta_0=1.0$, the initial step size; $\sigma=10^{-3}$,~the \lzero-norm smoothing hyperparameter; $\tau_0=0.3$,~the initial sparsity threshold; $t=0.01$, the sparsity threshold adjustment factor.}
  \KwOutput{$\vct x^\star$, the minimum \lzero-norm adversarial example.}
  $\boldsymbol{\delta} \leftarrow \vct 0$; $\quad \vct\delta^\star \leftarrow \vct \infty$; $\quad \tau \leftarrow \tau_0$; $\quad \eta \leftarrow \eta_0$  \label{line:initialization}


    \For{i {\rm in} $1, \dots, N$} {\label{line:sparsity_minimization}

        $\nabla\vct{g} \leftarrow \nabla_{\vct\delta}[\set{L}(\vct x + \vct \delta, y, \vct \theta) + \frac{1}{d}\lzerohat(\vct\delta, \sigma)]$\label{line:objective_gradient} \tcp*{Gradient Descent for \Cref{eq:attack-obejctive-approximated}.}
         
        $\nabla\vct{g} \leftarrow {\nabla\vct{g}}/{\|\nabla\vct{g}\|_\infty}$\label{line:gradient_normalization} \tcp*{Gradient Normalization.}

        $\vct\delta \leftarrow$ \texttt{clip}$(\vct x + [\vct\delta - \eta\cdot\nabla\vct{g}]) - \vct x$~\label{line:final_update} \tcp*{Box Constraints.}

        $\vct\delta \leftarrow \Pi_\tau(\vct\delta)$\label{line:tau_threshold_enforcement}  \tcp*{Adaptive Projection Operator.}
        
        $\eta = \texttt{cosine\_annealing}(\eta_0, i)$\label{line:cosinescheduler} \tcp*{Learning Rate Decay.}

        \rm \textbf{if} {$\set{L}(\vct x + \vct \delta, y, \vct \theta) \leq 0$}:
        { $\tau += t\cdot \eta$, {\rm \textbf{else} $\tau -= t\cdot \eta$}\label{line:tau_threshold_increase}} \tcp*{Adaptive Adjustment for $\tau$.}

        \rm \textbf{if} {$\set{L}(\vct x + \vct \delta, y, \vct \theta) \leq 0 ~\wedge ~ \|\vct\delta\|_0< \|\vct\delta^\star\|_0$}: 
        {  $\vct\delta^\star \leftarrow \vct\delta$\label{line:best_delta}}
        }

        \rm \textbf{if} {$\set{L}(\vct x + \vct \delta^\star, y, \vct \theta) > 0$:}{ { $\vct\delta^\star \leftarrow \infty$\label{line:best_delta_infty}}
        }

    \KwRet{$\vct x^\star \leftarrow \vct x + \vct \delta^\star$}\label{line:end_attack}
\caption{\sigmazero Attack Algorithm.}
\label{alg:sigma-zero}
\end{algorithm}

\myparagraph{Remarks.} To summarize, the main contributions behind \sigmazero are: 
(i) the use of a smooth \lzero-norm approximation, along with the definition of an appropriate objective (Eq.~\ref{eq:attack-obejctive-approximated}), to enable optimizing \lzero-norm adversarial examples via gradient descent; and (ii) the introduction of an adaptive projection operator to further improve sparsity during the optimization. 
Our algorithm leverages also common strategies like gradient normalization and step size annealing to speed up convergence. 
As reported by our experiments, \sigmazero provides a more effective and efficient $\ell_0$-norm attack that 
(i)~is robust to different hyperparameter choices; (ii)~does not require any adversarial initialization; and (iii)~enables more reliable robustness evaluations, being able to find adversarial examples also when the competing attacks may fail~\citep{Carlini2019EvalRobustness,pintor2022indicators}.


\section{Experiments}\label{sec:experiments}
We report here an extensive experimental evaluation comparing \sigmazero against 11 state-of-the-art sparse attacks, including both \lzero- and \lone-norm attacks. We test all attacks using different settings on 18 distinct models and 3 different datasets, yielding almost 450 different comparisons in total.
\subsection{Experimental Setup}\label{sec:experimental_setup}
\myparagraph{Datasets.} We consider the three most popular datasets used for benchmarking adversarial robustness: MNIST~\citep{LeCun2005TheMD}, CIFAR-10~\citep{Krizhevsky2009LearningML} and ImageNet~\citep{Krizhevsky2012ImageNetCW}. 
To evaluate the attack performance, 
we use the entire test set for MNIST and CIFAR-10 (with a batch size of 32), and a subset of $1000$ test samples for ImageNet (with a batch size of 16). 

\myparagraph{Models.} We use a selection of both baseline and robust models to evaluate the attacks under different conditions. 
We evaluate \sigmazero on a vast set of models to ensure its broad effectiveness and expose vulnerabilities that may not be revealed by other attacks~\citep{Croce2021MindTheBox}.
For the MNIST dataset, we consider two adversarially trained convolutional neural network (CNN) models by \citet{Rony21Alma}, i.e., \smallcnnddn and \smallcnntrades. These models have been trained to be robust to both \ltwo and \linf adversarial attacks. We denote them \smallcnnddnshort and \smallcnntradesshort, respectively.
For the CIFAR-10 and ImageNet datasets, we employ state-of-the-art robust models from RobustBench~\citep{Croce2021Robustbench} and the paper's official repositories. 
For CIFAR-10, we adopt ten models, denoted as C1-C12.  \carmonshort~\carmon and \augustinshort~\augustin combine training data augmentation with adversarial training to improve robustness to \linf and \ltwo attacks. 
\croceloneshort~\crocelone and \jiangloneshort~\jianglone are \lone robust models.  
\standardshort~\standard is a non-robust WideResNet-28-10 model.
\gowalshort~\gowal uses generative models to artificially augment the original training set and improve adversarial robustness to generic $\ell_p$-norm attacks.  
\engstromshort~\engstrom is an adversarially trained model that is robust against \ltwo-norm attacks.
\chenshort~\chen is a robust ensemble model. \xushort~\xu is a recently proposed adversarial training defense robust to \ltwo attacks. 
\addepallishort~\addepalli enforces diversity during data augmentation and combines it with adversarial training.
Lastly, \zhongtradesmed and \zhongsatmed are two adversarially trained models robust against \lzero-norm adversarial perturbations. 
For ImageNet, we consider a pretrained ResNet-18 denoted with \resnetvulnerableshort~\resnetvulnerable, and five robust models to \linf-attacks, denoted with \engstromimagenetshort~\engstromimagenet, \hendrycksshort~\hendrycks, \debenedettishort~\debenedetti, \wongshort~\wong, and \salmanshort~\salman.
\rebuttal{Lastly, in the appendix, we present two \lzero-robust models, \zhongtradesmed and \zhongsatmed, for CIFAR-10, along with two large \linf-robust models, \pengmed and \momed, for ImageNet.}

\myparagraph{Attacks.} We compare \sigmazero against the following state-of-the-art minimum-norm attacks, in their \lzero-norm variants: the Voting Folded Gaussian Attack (VFGA) attack~\citep{Cesaire2021VFGA}, the Primal-Dual Proximal Gradient Descent (PDPGD) attack~\citep{Matyasko2021PDPGD}, the Brendel \& Bethge (BB) attack~\citep{Brendel2019BB}, including also its variant with adversarial initialization (BBadv),\footnote{We utilize the Foolbox DatasetAttack~\citep{FoolboxDatasetAttack} for adversarial initialization.} and the Fast Minimum Norm (FMN) attack~\citep{Pintor2021FMN}. 
We also consider two state-of-the-art \lone-norm attacks as additional baselines, i.e., the Elastic-Net (EAD) attack~\citep{Chen2018EAD} and SparseFool (SF) by~\citet{modas-19-sparsefool}.
All attacks are set to manipulate the input values independently; 
e.g., for CIFAR-10, the number of modifiable inputs is $3\times 32 \times 32=3072$.

\myparagraph{Hyperparameters.} We run our experiments using the default hyperparameters from the original implementations provided in the authors' repositories,  \textit{AdversarialLib}~\citep{Rony_Adversarial_Library} and \textit{Foolbox}~\citep{Foolbox}. 
We set the maximum number of iterations to $N=1000$ to ensure that all attacks reach convergence~\citep{pintor2022indicators}.\footnote{Additional results using only $N=100$ steps are reported in Appendix~\ref{appendix:hundred_steps}.} 
For \sigmazero, we set $\eta_0=1$, $\tau_0=0.3$, $t=0.01$, and $\sigma=10^{-3}$, and keep the same configuration for all models and datasets.\footnote{To show that no specific hyperparameter tuning is required, additional results are reported in Appendix~\ref{appendix:ablation}.}

\myparagraph{Evaluation Metrics.} For each attack, we report the Attack Success Rate (ASR) at different values of $k$, denoted with ASR$_k$, i.e., the fraction of successful attacks for which $\|\vct \delta^\star\|_0 \leq k$, and the median value of $\|\vct \delta^\star \|_0$ over the test samples, denoted with $\tilde{\ell}_0$.\footnote{If no adversarial example is found for a given $\vct x$, we set  $\|\vct \delta^\star\|_0 = \infty$, as done by~\citet{Brendel2019BB}.} 
We compare the computational effort of each attack considering the mean runtime  (\textbf{s}) (per sample), the mean number of queries (\textbf{q}) (i.e., the total number of forwards and backwards required to perform the attack, divided by the number of samples), and the Video Random Access Memory (VRAM) consumed by the Graphics Processing Unit (GPU). 
We measure the runtime on a workstation with an NVIDIA A100 Tensor Core GPU (40 GB memory) and two Intel\textsuperscript{\textregistered} Xeo\textsuperscript{\textregistered} Gold 6238R processors. 
We evaluate memory consumption as the maximum VRAM used among all batches, representing the minimum requirement to run without failure.

\subsection{Experimental Results}\label{sec:experimental_results}
We report the success rate and computational effort metrics of \sigmazero against minimum-norm attacks in \autoref{tab:1k_minimum_norm_complete} and fixed-budget attacks in \autoref{tab:cifar-fixed_sigma-k}-\ref{tab:imagenet-fixed_sigma-k}. 
In these tables, we consider the most robust models for each dataset, and we provide the remaining results in Appendix~\ref{appendix:additional_experiments}. 
Finally, for ImageNet, we narrow our analysis to EAD, FMN, BBadv, and VFGA minimum-norm attacks, as they surpass competing attacks on MNIST and CIFAR-10 in terms of ASR, perturbation size, or execution time.
\begin{table*}[t]
\caption{Minimum-norm comparison results on MNIST, CIFAR10 and ImageNet with $N=1000$. For each attack and model ({M}), we report ASR at $k=24, 50, \infty$, median perturbation size \medianlzero, mean runtime $s$ (in seconds), mean number of queries $q$ (in thousands), and maximum VRAM usage (in GB). When VFGA exceeds the VRAM limit, we re-run it using a smaller batch size, increasing its runtime $t$. We denote those cases with the symbol `$\star$'. Remaining models in Appendix~\ref{appendix:additional_experiments}, \autoref{tab:1k_remaining_minimum_norm_cifar_imagenet}.}
\label{tab:1k_minimum_norm_complete}
\centering
\resizebox{1.0\textwidth}{!}{%
  \setlength\tabcolsep{1.5pt} 
\setlength{\dashlinedash}{5pt}
\setlength{\dashlinegap}{3pt}
\begin{tabular}{r|c|cccc|ccc|c|cccc|ccc}

\toprule
\multicolumn{1}{c}{\textbf{Attack}} & \multicolumn{1}{c}{\textbf{M}}& \multicolumn{1}{c}{\textbf{ASR$_{24}$}} & \multicolumn{1}{c}{\textbf{ASR$_{50}$}} & \multicolumn{1}{c}{\textbf{ASR$_\infty$}} & \multicolumn{1}{c}{\medianlzero} & \multicolumn{1}{c}{$\mathbf{s}$} & \multicolumn{1}{c}{$\mathbf{q}$}  & \multicolumn{1}{c}{\textbf{VRAM}} & \multicolumn{1}{c}{\textbf{M}} & \multicolumn{1}{c}{\textbf{ASR$_{24}$}} & \multicolumn{1}{c}{\textbf{ASR$_{50}$}} & \multicolumn{1}{c}{\textbf{ASR$_\infty$}} & \multicolumn{1}{c}{\medianlzero} & \multicolumn{1}{c}{$\mathbf{s}$} & \multicolumn{1}{c}{$\mathbf{q}$}  & \multicolumn{1}{c}{\textbf{VRAM}}\\

\hline\hline
\multicolumn{17}{c}{MNIST} \\ \hline\hline

{SF} &  & 6.66 & 6.76 & 96.98 & 469 & 1.07 & 0.18 & 0.06 &  & 1.03 & 1.21 & 91.68 & 463 & 2.87 & 0.86 & 0.07 \\
EAD &  & 3.83 & 53.66 & 100.0 & 49 & 0.47 & 6.28 & 0.05 &  & 2.13 & 55.57 & 100.0 & 48 & 0.50 & 6.73 & 0.05 \\
{PDPGD} &  & 26.77 & 74.08 & 100.0 & 38 & 0.23 & 2.00 & 0.04 &  & 16.91 & 66.30 & 100.0 & 42 & 0.23 & 2.00 & 0.04 \\
{VFGA} &  & 43.58 & 82.42 & 99.98 & 27 & 0.05 & 0.77 & 0.21 &  & 5.00 & 39.33 & 99.95 & 57 & 0.05 & 1.33 & 0.21 \\
{FMN} &  & 35.90 & 93.74 & 100.0 & 29 & 0.21 & 2.00 & 0.04 &  & 50.74 & 91.84 & 99.41 & 24 & 0.22 & 2.00 & 0.04 \\
{BB} &  & 71.23 & 97.86 & 100.0 & 18 & 0.90 & 2.99 & 0.05 &  & 56.53 & 91.62 & 100.0 & 18 & 0.74 & 3.71 & 0.05 \\
{BBadv} &  & 67.06 & 91.23 & 100.0 & 19 & 0.77 & 2.01 & 0.07 &  & 29.17 & 40.88 & 100.0 & 89 & 0.71 & 2.01 & 0.07 \\
\cdashline{1-17}
\rc {\sigmazero} & \multirow{-8}{*}{\textit{\smallcnnddnshort}} & \textbf{83.79} & \textbf{99.98} & 100.0 & \textbf{16} & 0.31 & 2.00 & 0.04 & \multirow{-8}{*}{\textit{\smallcnntradesshort}} & \textbf{98.03} & \textbf{100.0} & 100.0 & \textbf{9} & 0.31 & 2.00 & 0.04 \\ 

\hline\hline
\multicolumn{17}{c}{CIFAR-10} \\ \hline\hline

{SF} &  & 18.71 & 18.77 & 56.39 & 3072 & 11.31 & 1.40 & 1.62 &  & 20.46 & 24.36 & 58.29 & 3072 & 1.63 & 0.48 & 0.66 \\
{EAD} &  & 16.32 & 30.38 & 100.0 & 90 & 1.92 & 5.70 & 1.47 &  & 13.01 & 13.23 & 100.0 & 800 & 0.94 & 4.89 & 0.65  \\
{PDPGD} &  & 26.84 & 42.50 & 100.0 & 63 & 0.64 & 2.00 & 1.32 &  & 22.30 & 35.13 & 100.0 & 75 & 0.41 & 2.00 & 0.59 \\
{VFGA} &  & 51.06 & 75.37 & 99.92 & 24 & 0.59 & 0.78 & 11.71 &  & 28.47 & 49.98 & 99.72 & 51 & 0.32 & 1.25 & 4.44 \\
{FMN} &  & 48.89 & 74.70 & 100.0 & 26 & 0.59 & 2.00 & 1.31 &  & 27.45 & 48.87 & 100.0 & 52 & 0.24 & 2.00 & 0.60  \\
{BB} &  & 13.27 & 14.24 & 14.70 & \binfty & 0.63 & 2.05 & 1.47 &  & 16.88 & 22.91 & 27.64 & \binfty & 1.04 & 2.25 & 0.65 \\
{BBadv} &  & 65.96 & 90.57 & 100.0 & 16 & 4.68 & 2.01 & 1.64 &  & 36.47 & 72.43 & 100.0 & 34 & 5.28 & 2.01 & 0.64  \\

\cdashline{1-17}

\rc{\sigmazero} & \multirow{-8}{*}{\textit \carmonshort} & \textbf{76.53} & \textbf{95.38} & 100.0 & \textbf{11} & 0.73 & 2.00 & 1.53 & \multirow{-8}{*}{\textit \croceloneshort} & \textbf{38.60} & \textbf{73.02} & \textbf{100.0} & \textbf{32} & 0.43 & 2.00 & 0.71 \\ \hline

{SF} &  & 19.66 & 21.22 & 98.74 & 3070 & 3.62 & 0.46 & 1.90 &  & 31.76 & 43.07 & 91.14 & 69 & 4.32 & 1.49 & 0.66 \\
{EAD} &  & 9.73 & 11.42 & 100.0 & 360 & 2.53 & 5.62 & 1.89 &  & 24.21 & 24.78 & 100.0 & 768 & 1.04 & 4.99 & 0.65 \\
{PDPGD} &  & 28.02 & 45.15 & 100.0 & 55 & 1.12 & 2.00 & 1.8 &  & 26.89 & 42.38 & 100.0 & 66 & 0.40 & 2.00 & 0.60 \\
{VFGA} &  & 39.58 & 66.50 & 99.62 & 34 & 0.48 & 0.94 & 16.53 &  & 46.71 & 69.47 & 99.83 & 28 & 0.25 & 0.82 & 4.22 \\
{FMN} &  & 39.30 & 71.70 & 100.0 & 33 & 1.08 & 2.00 & 1.8 &  & 43.06 & 62.96 & 100.0 & 34 & 0.35 & 2.00 & 0.59 \\
{BB} &  & 38.73 & 56.78 & 58.64 & 33 & 2.31 & 2.89 & 1.89 &  & 25.95 & 27.98 & 29.50 & \binfty & 0.54 & 2.09 & 0.65 \\
{BBadv} &  & 70.07 & 96.31 & 100.0 & 17 & 3.92 & 2.01 & 1.99& & 53.17 & 82.46 & 100.0 & 22 & 3.03 & 2.01 & 0.65 \\

\cdashline{1-17}

\rc{\sigmazero} & \multirow{-8}{*}{\textit \augustinshort} & \textbf{74.63} & \textbf{97.55} & 100.0 & \textbf{15} & 1.41 & 2.00 & 1.92 & \multirow{-8}{*}{\textit \jiangloneshort} & \textbf{55.42} & \textbf{82.92} & 100.0 & \textbf{20} & 0.42 & 2.00 & 0.72 \\

\hline\hline
\multicolumn{17}{c}{ImageNet} \\ \hline\hline

EAD &  & 35.4 & 36.3 & 100.0 & 460 & 4.13 & 2.69 & 0.46 &  & 27.0 & 28.4 & 100.0 & 981 & 19.25 & 5.49 & 1.41 \\
{VFGA} &  & 57.9 & 72.5 & 99.9 & 14 & 1.22$^\star$ & 1.08 & \minfty  &  & 46.7 & 59.5 & 97.9 & 31 & 6.93$^\star$ & 1.98 & \minfty \\
{FMN} &  & 62.6 & 81.0 & 100.0 & 12 & 0.73 & 2.00 & 0.66 &  & 49.1 & 67.7 & 100.0 & 25 & 1.98 & 2.00 & 2.30 \\
{BBadv} &  & 77.5 & 93.2 & 100.0 & 7 & 231.67 & 2.01 & 0.72 &  & 64.7 & 85.5 & 100.0 & 14 & 205.11 & 2.01 & 2.41 \\

\cdashline{1-17}

\rc{\sigmazero} & \multirow{-5}{*}{\textit \resnetvulnerableshort} & \textbf{82.6} & \textbf{95.9} & 100.0 & \textbf{5} & 1.18 & 2.00 & 0.84 & \multirow{-5}{*}{\textit \hendrycksshort} & \textbf{66.7} & \textbf{86.9} & 100.0 & \textbf{13} & 2.76 & 2.00 & 2.52 \\ \hline

{EAD} &  & 46.8 & 51.0 & 100.0 & 42 & 18.10 & 5.45 & 1.42 &  & 32.8 & 33.5 & 100.0 & 572 & 11.43 & 5.34 & 1.68 \\
{VFGA} &  & 54.7 & 63.4 & 96.7 & 12 & 8.21$^\star$ & 2.35 & \minfty &  & 40.0 & 46.5 & 95.5 & 66 & 33.88$^\star$ & 3.97 & \minfty \\
{FMN} &  & 57.8 & 67.0 & 100.0 & 9 & 1.97 & 2.00 & 2.30 &  & 40.3 & 47.2 & 100.0 & 58 & 4.28 & 2.00 & 2.97 \\
{BBadv} &  & 71.0 & 82.3 & 100 & 4 & 182.65 & 2.01 & 2.40 &  & 46.8 & 59..8 & 100.0 & 31 & 178.06 & 2.01 & 3.07 \\

\cdashline{1-17}

\rc{\sigmazero} & \multirow{-5}{*}{\textit \engstromimagenetshort} & \textbf{76.9} & \textbf{87.4} & 100.0 & \textbf{3} & 2.75 & 2.00 & 2.52 & \multirow{-5}{*}{\textit \debenedettishort} & \textbf{50.7} & \textbf{65.1} & 100.0 & \textbf{23} & 5.72 & 2.00 & 3.20 \\

\bottomrule
\end{tabular}
}
\end{table*}

\myparagraph{Effectiveness.} The median values of $||\vct{\delta}^\star||_0$, denoted as $\tilde{\ell}_0$, and the ASRs are reported in \autoref{tab:1k_minimum_norm_complete} for all models and datasets. 
To facilitate comparison, the attacks are sorted from the least to the most effective, on average.
In all dataset-model configurations, \sigmazero significantly outperforms all the considered attacks. 
Taking the best-performing attack among the fastest competitors as a reference (i.e., FMN), \sigmazero is able to find smaller perturbations and higher ASRs in all configurations. 
In particular, on CIFAR-10, \sigmazero reduces the median number of manipulated features from $52$ to $32$ against the most robust model (\croceloneshort), with an average reduction of $49\%$ across all models. 
On ImageNet, this improvement is even more pronounced, with a reduction of up to $58\%$. In the best case (\debenedettishort), the median $\|\vct{\delta}^\star\|_0$ is reduced from $58$ to $23$, and in the worst case (\engstromimagenetshort), from 9 to 3. 
Alternatively, the most competitive attack in finding small perturbations is BBadv, which is significantly slower and requires starting from an already-adversarial input. 
The ASR$_{\infty}$ of BB (i.e., without adversarial initialization) indeed decreases with increasing input dimensionality (e.g., CIFAR-10). This occurs because BB often stops unexpectedly before reaching the specified number of steps due to initialization failures; in particular, \autoref{tab:1k_minimum_norm_complete} shows that the median perturbation size found by BB is sometimes $\infty$, as its ASR$_{\infty}$ is lower than $50\%$.
Although BBadv does not suffer from the same issue, as it leverages adversarial initialization, it is still outperformed by \sigmazero. Specifically, \sigmazero reduces the $\ell_0$ norm of the adversarial examples from $16$ to $11$ in the best case~(\carmonshort), while achieving an average improvement of $24\%$ across all dataset-model configurations. 

\begin{figure*}[tb]
    \centering
    \includegraphics[width=1\textwidth]{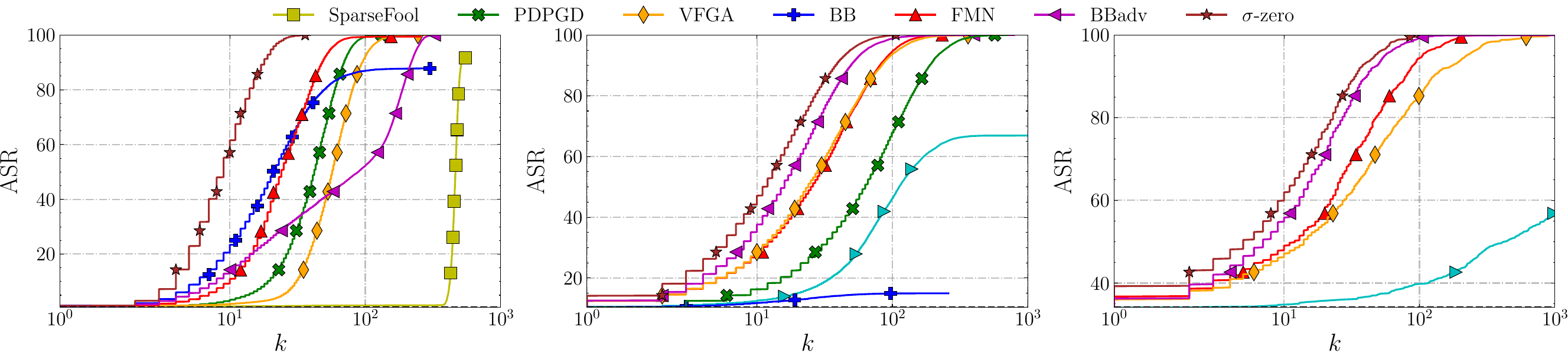}\hfill
    \caption{Robustness evaluation curves (ASR vs.  perturbation budget $k$) for \smallcnntradesshort on MNIST (\textit{left}), \carmonshort on CIFAR-10 (\textit{middle}), and \resnetvulnerableshort on ImageNet (\textit{right}). 
    }
    \label{fig:sec_curves}
\end{figure*}
\myparagraph{Efficiency.} We evaluate the computational effort required to run each attack by reporting in \autoref{tab:1k_minimum_norm_complete} the mean runtime $s$ (in seconds), the mean number of queries $q$ issued to the model (in thousands), and the maximum VRAM used. 
Note that, while the runtime $s$ and the consumed VRAM may depend on the attack implementation, the number of queries $q$ counts the total number of forward and backward passes performed by the attack, thus providing a fairer evaluation of the attack complexity. 
In fact, some attacks perform more than $2000$ queries even if $N=1000$, i.e., they perform more than one forward and one backward pass per iteration (see, e.g., EAD and BB). 
Other attacks, instead, might use less than $2000$ queries as they implement early stopping strategies.
The results indicate that \sigmazero exhibits similar runtime performance when compared to the fastest algorithms FMN, PDPGD, and VFGA, while preserving higher effectiveness. 
In contrast, when compared against the BBadv attack, which competes in terms of $\tilde{\ell}_0$, our attack is much faster across all the dataset-model configurations, especially for Imagenet.
For example, \sigmazero is 10 times faster than BBadv on \jiangloneshort and 100 times faster on \hendrycksshort on ImageNet. This confirms that \sigmazero establishes a better effectiveness-efficiency trade-off than that provided by state-of-the-art $\ell_0$-norm attacks.

\myparagraph{Reliability.} Complementary to \autoref{tab:1k_minimum_norm_complete}, we present the robustness evaluation curves in \autoref{fig:sec_curves} for each attack on \smallcnntradesshort, \carmonshort, and \resnetvulnerableshort.
In Appendix~\ref{appendix:sec_curves}, we include similar curves for all other configurations. 
These curves go beyond the only median statistic and ASR$_{k}$, providing further evidence that \sigmazero achieves higher ASRs with smaller \lzero-norm perturbations compared to the competing attacks. 
More importantly, the ASR of \sigmazero reaches almost always $100\%$ as the perturbation budget grows, meaning that its optimization only rarely fails to find an adversarial example. 
In Appendix~\ref{appendix:hundred_steps}, we further demonstrate that even when the number of iterations is reduced to $N=100$, \sigmazero consistently achieves an ASR$_{\infty}$ of $100\%$ across all models. This is not observed with other attacks, which often fail when using fewer iterations, thereby increasing the risk of overestimating adversarial robustness.
These results reinforce our previous findings, confirming that \sigmazero can help mitigate the issue of overestimating adversarial robustness -- a crucial aspect to foster scientific progress in defense developments and evaluations ~\citep{Carlini2019EvalRobustness,pintor2022indicators}.

\myparagraph{Ablation Study.} In \autoref{tab:ablation_components_table} we present an ablation study to evaluate the relevance of \sigmazero's components. 
Our findings indicate that all the non-trivial components in \sigmazero are essential for ensuring the effectiveness of the attack.
Specifically, we observe that the \lzero-norm approximation $\hat{\ell}_0$ (Eq.~\ref{eq:l0_approximation}, \autoref{line:objective_gradient}) leads the optimization algorithm to perturb all input features, albeit with small contributions. 
The projection operator (\autoref{line:tau_threshold_enforcement}) plays a crucial role by significantly decreasing the number of perturbed features, effectively removing the least significant contributions. 
Furthermore, gradient normalization (\autoref{line:gradient_normalization})  accelerates convergence, enhancing efficiency. 
Lastly, the adaptive projection operator (\autoref{line:tau_threshold_increase}) fine-tunes the results, reduces the number of perturbed features, and mitigates the dependency on hyperparameter choices.
These results underline the importance of each component in \sigmazero, highlighting their contributions to the overall performance of the attack.

\begin{table}[htbp]
\caption{Ablation study on the \sigmazero components integrated in \autoref{alg:sigma-zero}. Columns describe respectively: Gradient normalization factor~(\autoref{line:gradient_normalization}); dynamic projection adjustment \autoref{line:tau_threshold_increase}; projection operator $\Pi_\tau$ (\autoref{line:tau_threshold_enforcement}); and the \lzero norm approximation $\hat{\ell}_0$ (\autoref{line:objective_gradient}).}
\label{tab:ablation_components_table}
\centering
\begin{tabular}{@{}cclcccccc@{}}
\toprule
Model & Normalization & \multicolumn{1}{c}{Adaptive $\tau$} & Projection & $\hat{\ell}_0$ & ASR$_{10}$ & ASR$_{50}$ & ASR & \medianlzero \\ \midrule
\multirow{6}{*}{C10} & \tcheck & \multicolumn{1}{c}{\tcheck} & \tcheck & \tcheck & 21.68 & \textbf{73.02} & 100.0 & \textbf{32} \\
 & \tcheck\cc &  \cc& \tcheck \cc& \tcheck \cc& \textbf{21.89} \cc& 71.66 \cc& 100.0 \cc& \textbf{32} \cc\\
 & \multicolumn{1}{l}{} & \multicolumn{1}{c}{\tcheck} & \tcheck & \tcheck & 16.81 & 39.76 & 100.0 & 65 \\
 & \cc & \cc & \tcheck \cc & \tcheck \cc & 12.95 \cc & 13.23 \cc & 100.0 \cc & 505 \cc\\
 & \multicolumn{1}{l}{} &  & \multicolumn{1}{l}{} & \tcheck & 12.95 & 12.95 & 100.0 & 3004 \\
 & \tcheck \cc &  \cc&  \cc & \tcheck \cc & 12.95 \cc & 12.95 \cc& 100.0 \cc & 3070 \cc \\

\midrule \midrule

\multirow{6}{*}{C5} & \tcheck & \multicolumn{1}{c}{\tcheck} & \tcheck & \tcheck & \textbf{37.27} & \textbf{82.92} & 100.0 & \textbf{20} \\
 & \tcheck \cc & \cc & \tcheck\cc & \tcheck \cc& 37.01 \cc & 79.83\cc & 100.0 \cc& 21\cc \\
 & \multicolumn{1}{l}{} & \multicolumn{1}{c}{\tcheck} & \tcheck & \tcheck & 29.56 & 52.83 & 100.0 & 46 \\
 & \cc & \cc & \tcheck \cc& \tcheck\cc & 25.46\cc & 32.84\cc & 100.0 \cc& 144 \cc\\
 & \multicolumn{1}{l}{} &  & \multicolumn{1}{l}{} & \tcheck & 23.78 & 23.78 & 100.0 & 3064 \\
 & \tcheck \cc& \cc & \cc & \tcheck \cc& 23.78 \cc& 23.78 \cc& 100.0 \cc& 3068 \cc\\ \bottomrule
\end{tabular}
\end{table}

\begin{table}[tb]
\centering
\caption{Fixed-budget comparison results with $N=1000$ ($N=2000$ for Sparse-RS) on MNIST and CIFAR-10 at budgets $k = 24, 50, 100$. Columns q$_{24}$ and s$_{24}$ show the average number of queries (in thousands) and the average execution time per sample (in seconds) at $k=24$.}
\setlength\tabcolsep{2.7pt} 
\label{tab:cifar-fixed_sigma-k}

\centering
\resizebox{1.0\textwidth}{!}{%
  \setlength\tabcolsep{1.5pt} 
\setlength{\dashlinedash}{5pt}
\setlength{\dashlinegap}{3pt}
\begin{tabular}{r|c|ccc|ccc|c|ccc|ccc}

\toprule
\multicolumn{1}{c}{\textbf{Attack}} & \multicolumn{1}{c}{\textbf{M}} & \multicolumn{1}{c}{\textbf{ASR}$_{24}$} & \multicolumn{1}{c}{\textbf{ASR}$_{50}$} & \multicolumn{1}{c}{\textbf{ASR}$_{\rm 100}$} & \multicolumn{1}{c}{\textbf{q}$_{24}$} & \multicolumn{1}{c}{$\textbf{s}_{\rm 24}$}&  \multicolumn{1}{c}{$\textbf{VRAM}$} & \multicolumn{1}{c}{\textbf{M}} & \multicolumn{1}{c}{\textbf{ASR}$_{24}$} & \multicolumn{1}{c}{\textbf{ASR}$_{50}$} & \multicolumn{1}{c}{\textbf{ASR}$_{\rm 100}$} & \multicolumn{1}{c}{\textbf{q}$_{\rm 24}$}  & \multicolumn{1}{c}{\textbf{s}$_{\rm 24}$} & \multicolumn{1}{c}{$\textbf{VRAM}$} \\

\hline\hline
\multicolumn{15}{c}{MNIST} \\ \hline\hline

\PGDlzero & \multirow{5}{*}{\smallcnnddnshort} & 73.99 & 99.90 & 100.0 & 2.00 & 0.09 & 0.04& \multirow{5}{*}{\smallcnntradesshort} & 61.87 & 94.15 & 98.50 & 2.00 & 0.09 & 0.04 \\
Sparse-RS &  & 79.54 & 96.35 & 99.79 & 0.83 & 0.21 & 0.04 & & \textbf{98.92} & 99.96 & 100.0 & 0.24 & 0.07 & 0.04 \\
\SPGDproj &  & 65.55 & 97.97 & 99.99 & 0.46 & 0.09 & 0.05 & & 67.92 & 98.57 & 99.97 & 0.92 & 0.08 & 0.05 \\
\SPGDunproj &  & 82.79 & 99.65 & 100.0 & 0.09 & 0.08 & 0.05 & & 62.25 & 98.11 & 99.99 & 1.00 & 0.09 & 0.05\\

\cdashline{1-15}

\rc\sigmazero &  & \textbf{83.71} & \textbf{99.98} & 100.0 & 0.43 & 0.02 & 0.06 & & 98.11 & \textbf{100.0} & 100.0 & 0.14 & 0.01 & 0.06 \\ 

\hline\hline
\multicolumn{15}{c}{CIFAR-10} \\ \hline\hline

\PGDlzero & \multirow{5}{*}{\carmonshort} & 38.18 & 59.67 & 87.19 & 2.00 & 0.78 & 1.90 & \multirow{5}{*}{\croceloneshort} & 22.99 & 36.20 & 67.54 & 2.00 & 0.35 & 0.69 \\
Sparse-RS &  & 72.51 & 86.59 & 94.28 & 0.77 & 0.36 & 1.95 & & 30.87 & 45.65 & 63.26 & 1.47 & 0.28 & 0.68 \\
\SPGDproj &  & 66.37 & 89.21 & 99.36 & 0.74 & 0.41 & 2.06 & & 31.82 & 58.62 & 93.19 & 1.39 & 0.17 & 0.73 \\
\SPGDunproj &  & 66.33 & 91.07 & 99.75 & 0.72 & 0.41 & 2.06 &  & 36.16 & 70.06 & 98.07 & 1.30 & 0.16 & 0.73\\

\cdashline{1-15}

\rc\sigmazero &  & \textbf{77.08} & \textbf{95.33} & \textbf{99.95} & 0.65 & 0.29 & 2.07 & & \textbf{38.67} & \textbf{73.00} & \textbf{98.53} & 1.33 & 0.15 & 0.75 \\
\hline

\PGDlzero & \multirow{5}{*}{\augustinshort} & 32.41 & 59.19 & 89.22 & 2.00 & 0.57 & 2.46 & \multirow{5}{*}{\jiangloneshort} & 34.35 & 44.99 & 68.61 & 2.00 & 0.35  & 0.70\\
Sparse-RS &  & 59.24 & 79.81 & 92.43 & 1.04 & 0.35 & 2.46 & & 49.35 & 63.01 & 76.51 & 1.11 & 0.37  & 0.68 \\
\SPGDproj &  & 58.91 & 88.15 & 99.42 & 0.89 & 0.39 & 2.57  & & 50.41 & 75.86 & 97.52 & 1.02 & 0.18 & 0.73\\
\SPGDunproj &  & 64.8 & 93.15 & 99.92 & 0.76 & 0.48 & 2.56 &  & \textbf{55.89} & \textbf{84.64} & \textbf{99.56} & 0.91 & 0.19 &  0.73 \\

\cdashline{1-15}

\rc\sigmazero &  & \textbf{75.09} & \textbf{97.67} & \textbf{100.0} & 0.65 & 0.17 & 2.68 &  & 55.69 & 82.72 & 99.07 & 0.94 & 0.11 & 0.75 \\ \bottomrule
\end{tabular}}
\end{table}
\begin{table}[t]
\centering
\caption{Fixed-budget comparison results with $N=1000$ ($N=2000$ for Sparse-RS) on ImageNet at budgets $k = 100,150$. See the caption of \autoref{tab:cifar-fixed_sigma-k} for further details.}
\label{tab:imagenet-fixed_sigma-k}
\centering
  \setlength\tabcolsep{2.7pt} 
\setlength{\dashlinedash}{5pt}
\setlength{\dashlinegap}{3pt}
\begin{tabular}{r|c|cc|ccc|c|cc|ccc}
\toprule
\multicolumn{1}{c}{\textbf{Attack}} & \multicolumn{1}{c}{\textbf{M}} & \multicolumn{1}{c}{\textbf{ASR}$_{100}$} & \multicolumn{1}{c}{\textbf{ASR}$_{150}$} & \multicolumn{1}{c}{\textbf{q}$_{100}$} &  \multicolumn{1}{c}{$\textbf{s}_{\rm 100}$} & \multicolumn{1}{c}{\textbf{VRAM}} & \multicolumn{1}{c}{\textbf{M}} & \multicolumn{1}{c}{\textbf{ASR}$_{100}$} & \multicolumn{1}{c}{\textbf{ASR}$_{150}$} & \multicolumn{1}{c}{\textbf{q}$_{100}$} &  \multicolumn{1}{c}{$\textbf{s}_{\rm 100}$} & \multicolumn{1}{c}{\textbf{VRAM}} \\ 

\hline\hline
\multicolumn{13}{c}{ImageNet} \\ \hline\hline

Sparse-RS & \multirow{4}{*}{\resnetvulnerableshort} & 89.3 & 91.5 & 0.39  & 0.32 & 1.29 &\multirow{4}{*}{\engstromimagenetshort} & 81.1 & 84.1 & 0.53 & 0.5 & 4.39 \\
\SPGDproj &  & 95.4 & 98.5 & 0.31 & 0.16 & 1.40 & & 85.6 & 91.2 & 0.33 & 0.64 & 4.48\\
\SPGDunproj &  & 93.6 & 97.8 & 0.33 & 0.12 & 1.40 & & 82.6 & 88.7 & 0.37 & 0.39 & 4.49\\
\cdashline{1-13}

\rc \sigmazero &  & \textbf{99.7} & \textbf{100.0} & 0.19 & 0.06 & 1.79 & & \textbf{94.7} & \textbf{97.1} & 0.15 & 0.17 & 4.90 \\

\hline

Sparse-RS & \multirow{4}{*}{\hendrycksshort} & 69.1 & 72.2 & 0.81 & 0.62 & 4.39 & \multirow{4}{*}{\debenedettishort} & 45.9 & 47.4 & 1.17 & 1.12 &5.72 \\
\SPGDproj &  & 85.4 & 93.4 & 0.32 & 0.55 & 4.49 & & 66.3 & 74.9 & 0.73 & 1.39 & 5.84 \\
\SPGDunproj &  & 83.9 & 92.1 & 0.35 & 0.39 & 4.49 & & 66.0 & 76.0 & 0.72 & 1.01 & 5.84 \\
\cdashline{1-13}

\rc \sigmazero &  & \textbf{97.7} & \textbf{99.6} & 0.34 & 0.37 & 4.90 & & \textbf{78.8} & \textbf{85.8} & 0.49 & 0.70 & 6.29 \\ \bottomrule
\end{tabular}
\end{table}

\myparagraph{Comparison with Fixed-budget Attacks.} 
We complement our analysis by comparing \sigmazero with three fixed-budget \lzero-norm attacks, i.e., the \lzero-norm Projected Gradient Descent (\PGDlzero) attack~\citep{Croce2019SparseAI}, the Sparse Random Search (Sparse-RS) attack~\citep{Croce22SparseRS},\footnote{Sparse-RS is a gradient-free (black-box) attack, which only requires query access to the target model. 
We consider it as an additional baseline in our experiments, but it should not be considered a direct competitor of gradient-based attacks, as it works under much stricter assumptions (i.e., no access to input gradients).} and the Sparse-PGD attack~\citep{zhong2024towards}.
For Sparse-PGD, we consider the implementation with sparse (\SPGDproj) and with unprojected (\SPGDunproj) gradient.
In contrast to minimum-norm attacks, fixed-budget attacks optimize adversarial examples within a given maximum perturbation budget $k$. 
For a fairer comparison,as done in fixed-budget approaches, we early stop the \sigmazero optimization process as soon as an adversarial example with an $\ell_0$-norm perturbation smaller than $k$ is found.
In these evaluations, we set $N=1000$ for \sigmazero, \PGDlzero, \SPGDproj, and \SPGDunproj, while using $N=2000$ for Sparse-RS. 
\rebuttal{Therefore, when using $N=1000$ steps for \sigmazero (which amounts to performing 1000 forward and 1000 backward calls), we set $N=2000$ steps for Sparse-RS (which amounts to performing $2000$ forward calls).\footnote{\rebuttal{$N=2000$ is suggested as a lower bound number of iterations to ensure the convergence of Sparse-RS by~\citet{Croce22SparseRS}. Additional results with $N=5000/10000$ for Sparse-RS can be found in Appendix~\ref{appendix:maxattacks}.}} }
Furthermore, to compute the ASR at different $k$ (ASR$_k$), we separately execute fixed-budget attacks for $k=24, 50, 100$ features on MNIST and CIFAR-10, and with $k=100,150$ features on ImageNet (excluding \PGDlzero due to computational demands), reporting only the maximum number of queries and execution time across all distinct runs.
We report the average query usage at $k$ (\textbf{q}$_k$) and the average execution time per sample at $k$ ($\textbf{s}_k$).
We report the execution time of $\textbf{s}_k$ for the smaller $k$, as it requires, on average, more iterations due to the more challenging problem.
The results, shown in Tables~\ref{tab:cifar-fixed_sigma-k}-\ref{tab:imagenet-fixed_sigma-k}, confirm that \sigmazero outperforms competing approaches in 17 out of 18 configurations (see Appendix~\ref{appendix:maxattacks} for additional results).
Only against \jiangloneshort the fixed-budget attack \SPGDunproj slightly increases the ASR.
The advantages of \sigmazero become even more evident when looking at the results on ImageNet, where, on average, it improves the ASR$_{100}$ of $9.6\%$ across all models in \autoref{tab:imagenet-fixed_sigma-k}. 
The results also indicate that early stopping enables \sigmazero to save a significant number of queries and runtime while preserving a high ASR. 
In Appendix~\ref{appendix:maxattacks}, we also report additional comparisons with $N=2500$ and $N=5000$, i.e. a more favorable scenario for the competing attacks, confirming that \sigmazero remains competitive even at higher budgets.

\myparagraph{Summary.}
Our experiments show that \sigmazero:
(i)~outperforms minimum-norm attacks by improving the success rate and decreasing the $\ell_0$ norm of the generated adversarial examples (see \autoref{tab:1k_minimum_norm_complete} and Appendix~\ref{appendix:hundred_steps}); (ii)~is significantly faster and scales easily to large datasets (see \autoref{tab:1k_minimum_norm_complete} and Appendix~\ref{appendix:hundred_steps}); (iii)~is robust to hyperparameter selection, not requiring sophisticated and time-consuming tuning (see Appendix~\ref{appendix:ablation}); (iv)~does not require any adversarial initialization (see \autoref{tab:1k_minimum_norm_complete});
(v)~provides more reliable adversarial robustness evaluations, consistently achieving $100\%$ ASRs (see \autoref{tab:1k_minimum_norm_complete}, \autoref{fig:sec_curves}, Appendix~\ref{appendix:sec_curves}); and (vi)~remains competitive against fixed-budget attacks even when given the same query budget (\autoref{tab:cifar-fixed_sigma-k}-\ref{tab:imagenet-fixed_sigma-k}). 

\section{Related Work}\label{sec:relatedwork}
Optimizing \lzero-norm adversarial examples with gradient-based algorithms is challenging due to non-convex and non-differentiable constraints. 
We categorize them into two main groups: (i)~multiple-norm attacks extended to \lzero, and (ii)~attacks specifically designed to optimize the \lzero norm. 

\myparagraph{Multiple-norm Attacks Extended to \lzero}.
These attacks have been developed to work with multiple \lp norms, including extensions for the \lzero norm. While they can find sparse perturbations, they often rely heavily on heuristics in this setting.
\citet{Brendel2019BB} initialize the attack from an adversarial example far away from the clean sample and optimizes the perturbation by following the decision boundary to get closer to the source sample. In general, the algorithm can be used for any \lp norm, including \lzero, but the individual optimization steps are very costly.
\citet{Pintor2021FMN} propose the FMN attack that does not require an initialization step and converges efficiently with lightweight gradient-descent steps. However, their approach was developed to generalize over \lp norms, but does not make special adaptations to minimize the \lzero norm specifically.
\citet{Matyasko2021PDPGD} 
use relaxations of the \lzero norm (e.g., $\ell_{1/2}$) to promote sparsity. However, this scheme does not strictly minimize the \lzero norm, as the relaxation does not set the lowest components exactly to zero. 

\myparagraph{$\boldsymbol{\ell}_0$-specific Attacks.}
\citet{Croce22SparseRS} introduced Sparse-RS, a random search-based attack that, unlike minimum-norm attacks, aims to find adversarial examples that are misclassified with high confidence within a fixed perturbation budget. 
On the same track we find Sparse-PGD~\citep{zhong2024towards} and \PGDlzero~\citep{Croce2019SparseAI}, white-box fixed-budget alternatives to Sparse-RS.
Lastly, \citet{Cesaire2021VFGA} 
induces folded Gaussian noise to selected input components, iteratively finding the set that achieves misclassification with minimal perturbation. However, it requires considerable memory to explore possible combinations and find an optimal solution, limiting its scalability.

Overall, current implementations of \lzero-norm attacks present a crucial suboptimal trade-off between their success rate and efficiency, i.e., they are either accurate but slow (e.g., BB) or fast but inaccurate (e.g., FMN). 
This is also confirmed by a recent work that has benchmarked more than $100$ gradient-based attacks~\citep{cina2024attackbench} on $9$ additional robust models. In that open-source benchmark, \sigmazero consistently and significantly outperformed all the existing implementations of competing \lzero-norm attacks, establishing a performance very close to that of the empirical \textit{oracle} (obtained by ensembling all the attacks tested).
In summary, our attack combines the benefits of the two families of attack detailed above, i.e., effectiveness and efficiency, providing the state-of-the-art solution for adversarial robustness evaluations of DNNs when considering $\ell_0$-norm attacks. 


\section{Conclusions and Future Work}
\label{sec:conclusion}
In this work, we propose \sigmazero, a novel attack aimed to find minimum $\ell_0$-norm adversarial examples, based on the following main technical contributions: (i) a differentiable approximation of the $\ell_0$ norm to define a novel, smooth objective that can be minimized via gradient descent; and (ii) an adaptive projection operator to enforce sparsity in the adversarial perturbation, by zeroing out the least relevant features in each iteration. \sigmazero also leverages specific optimization tricks to stabilize and speed up the optimization.
Our extensive experiments demonstrate that \sigmazero consistently discovers more effective and reliable $\ell_0$-norm adversarial perturbations across all models and datasets while maintaining computational efficiency and robustness to hyperparameters choice.
In conclusion, \sigmazero emerges as a highly promising candidate to evaluate robustness against \lzero-norm perturbations and promote the development of novel robust models against sparse attacks. 

\myparagraph{Ethics Statement.} 
Based on our comprehensive analysis, we assert that there are no identifiable ethical considerations or foreseeable negative societal consequences that warrant specific attention within the limits of this study. This study will rather help improve the understanding of adversarial robustness of DNNs and identify potential ways to improve it.

\myparagraph{Reproducibility.} 
To ensure the reproducibility of our work, we have detailed the experimental setup in \Cref{sec:experimental_setup}, where we describe the datasets, models, and attacks used, along with their respective sources. Additionally, we have provided our source code as part of the supplementary material, which will be made publicly available as open source upon acceptance. 

\subsubsection*{Acknowledgments}
This work has been partially supported by 
the project Sec4AI4Sec, under the EU’s Horizon Europe Research and Innovation Programme (grant agreement no. 101120393); 
the project ELSA, under the EU’s Horizon Europe Research and Innovation Programme (grant agreement no. 101070617); 
the EU—NGEU National Sustainable Mobility Center (CN00000023), Italian Ministry of University and Research (MUR) Decree n. 1033—17/06/2022 (Spoke 10); 
projects SERICS (PE00000014) and FAIR (PE0000013) under the MUR NRRP funded by the EU—NGEU; 
and by the German Federal Ministry of Education and Research under the grant AIgenCY (16KIS2012).

\clearpage
\bibliography{preprintbib}
\bibliographystyle{preprintbib}

\clearpage
\appendix
\section{Appendix}
\subsection{Robust Models} 
The experimental setup described in this paper (Section~\ref{sec:experimental_setup}) utilizes pre-trained baseline and robust models obtained from RobustBench~\citep{Croce2021Robustbench}. The goal of RobustBench is to track the progress in adversarial robustness for \linf- and \ltwo-norm attacks since these are the most studied settings in the literature. We summarize in \autoref{tab:models_summary} the models we employed for testing the performance of \sigmazero. Each entry in the table includes the label reference from RobustBench, the short name we assigned to the model, and the corresponding clean and robust accuracy under the specific threat model. The robustness of these models is evaluated against an ensemble of white-box and black-box attacks, specifically AutoAttack.
We also include in our experiments models trained to be robust against \lone-sparse attacks, i.e., \croceloneshort~\crocelone and \jiangloneshort~\jianglone, as well as two robust models trained to resist \lzero-norm attacks, i.e., \zhongtradesmed~and \zhongsatmed. 
Our experimental setup is designed to encompass a wide range of model architectures and defensive techniques, ensuring a comprehensive and thorough performance evaluation of the considered attacks.

\begin{table}[htbp]
    \caption{Summary of Robustbench models used in our experiments. For each model, we report its reference label in Robustbench, its threat model, and the corresponding clean and robust accuracy.}
    \label{tab:models_summary}
    \resizebox{\textwidth}{!}{%
    \renewcommand{\arraystretch}{1.1}
    \begin{tabular}{@{}lllccc@{}}
    \toprule
    \textbf{Dataset} & \textbf{Reference} & \textbf{Model} & \textbf{Threat model} & \textbf{Clean accuracy} \% & \textbf{Robust accuracy} \% \\ \midrule
    \multirow{8}{*}{CIFAR-10} & Carmon2019Unlabeled & \carmonmed & \linf & 89.69 & 59.53 \\
     & Augustin2020Adversarial & \augustinmed & \ltwo & 91.08 & 72.91 \\
     & Standard & \standardmed & - & 94.78 & 0 \\
     & Gowal2020Uncovering & \gowalmed & \ltwo & 90.90 & 74.50 \\
     & Engstrom2019Robustness & \engstrommed & \linf - \ltwo & 87.03 - 90.83 & 49.25 - 69.24 \\
     & Chen2020Adversarial & \chenmed & \linf & 86.04 & 51.56 \\
     & Xu2023Exploring\_WRN-28-10 & \xumed & \linf & 93.69 & 63.89 \\
     & Addepalli2022Efficient\_RN18 & \addepallimed & \linf & 85.71 & 52.48
     \\ \midrule
    \multirow{6}{*}{ImageNet} & Standard\_R18 & \resnetvulnerablemed & - & 76.52 & 0 \\
     & Engstrom2019Robustness & \engstromimagenetmed & \linf & 62.56 & 29.22 \\
     & Hendrycks2020Many & \hendrycksmed & \linf & 76.86 & 52.90 \\
     & Debenedetti2022Light\_XCiT-S12 & \debenedettimed & \linf & 72.34 & 41.78 \\
     & Wong2020Fast & \wongmed & \linf & 55.62 & 26.24 \\
     & Salman2020Do\_R18 & \salmanmed & \linf & 64.02 & 34.96 \\
     & \rebuttal{Peng2023Robust} & \pengmed & \linf & 73.44 & 48.94 \\
     & \rebuttal{Mo2022When\_Swin-B} & \momed & \linf & 74.66 & 38.30  \\ \bottomrule
    \end{tabular}
    }
    \end{table}

\subsection{\sigmazero: Hyperparameter Robustness}\label{appendix:ablation}
To assess the strength and potential limitations of our proposed attack, we conducted an ablation study on its key hyperparameters, $\tau_0$, $\sigma$, and $t$.

\begin{figure*}[t]
    \centering
    \includegraphics[trim={0 1cm 0cm 0},clip,width=0.30\linewidth]{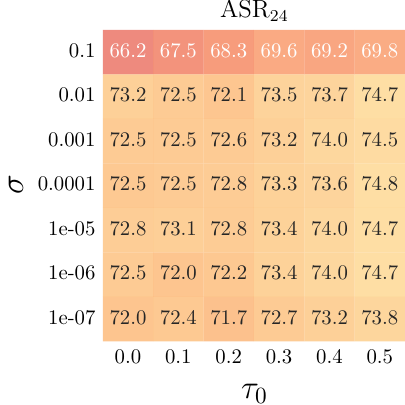}
    \includegraphics[trim={1.7cm 1cm 0cm 0},clip,width=0.226\linewidth]{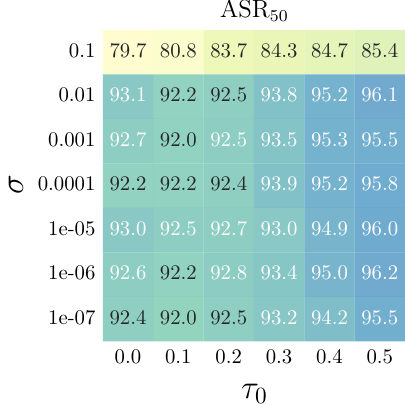}
    \includegraphics[trim={1.7cm 1cm 0cm 0},clip,width=0.226\linewidth]{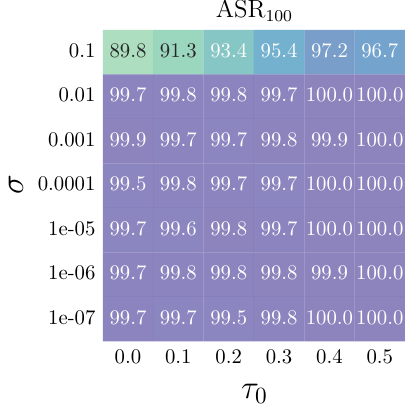}
    \includegraphics[trim={1.7cm 1cm 0cm 0},clip,width=0.229\linewidth]{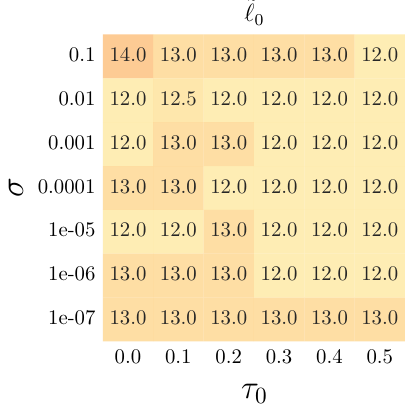}

    \includegraphics[trim={0 0cm 0cm 0.4cm},clip,width=0.30\linewidth]{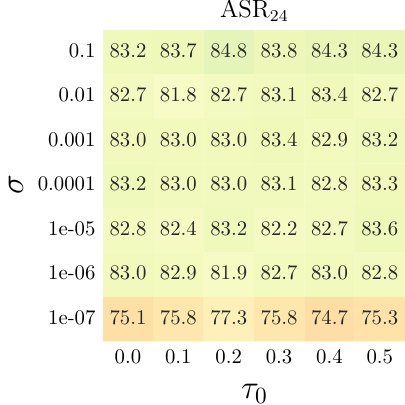}
    \includegraphics[trim={1.7cm 0cm 0cm 0.4cm},clip,width=0.226\linewidth]{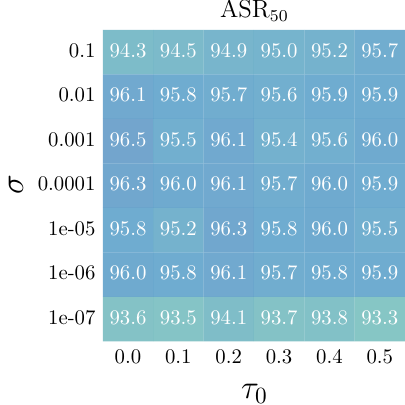}
    \includegraphics[trim={1.7cm 0cm 0cm 0.4cm},clip,width=0.226\linewidth]{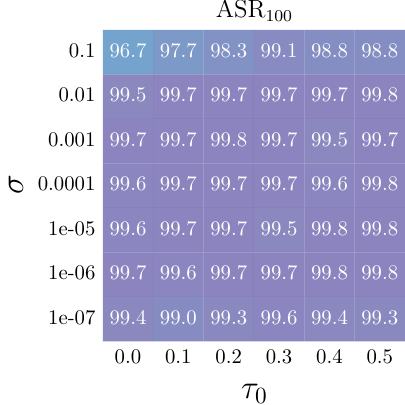}
    \includegraphics[trim={1.7cm 0cm 0cm 0.4cm},clip,width=0.229\linewidth]{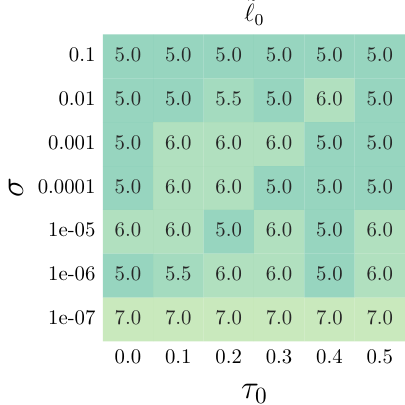}\hfill
    \caption{\small Ablation study on  $\sigma$ (y-axis) and $\tau_0$ (x-axis) for CIFAR-10 \addepallishort (top-row), ImageNet \resnetvulnerableshort, (bottom-row). For each combination, we report the attack success rate at different $k$ and the median $\ell_0$ perturbation value.}
    \label{fig:hyper}
\end{figure*}

The parameter $\tau_0$ governs the initial tolerance threshold in \autoref{alg:sigma-zero}, which induces sparsity within the adversarial perturbation. 
The parameter $\sigma$ defines the approximation quality of \lzerohat in \Cref{eq:l0_approximation} compared to the actual \lzero function.
Our ablation study, depicted in \autoref{fig:hyper}, involved two distinct models: \addepallishort{} (top row) and \resnetvulnerableshort (bottom row). 
We executed the attack on $1000$ randomly selected samples from each dataset and recorded the ASR at different perturbation budgets $k$ and the median \lzero norm of the resulting adversarial perturbations.
We observe a significant robustness of \sigmazero with respect to these two hyperparameters; in particular: (i) the choice of the initial value of $\tau_0$ exerts negligible influence on the ultimate outcome, given that the parameter dynamically adapts throughout the optimization process; and (ii) 
the selection of $\sigma$ is not particularly challenging, especially when incorporating the sparsity projection operator.

\begin{figure*}[!htb]
    \centering
    \includegraphics[width=0.45\textwidth]{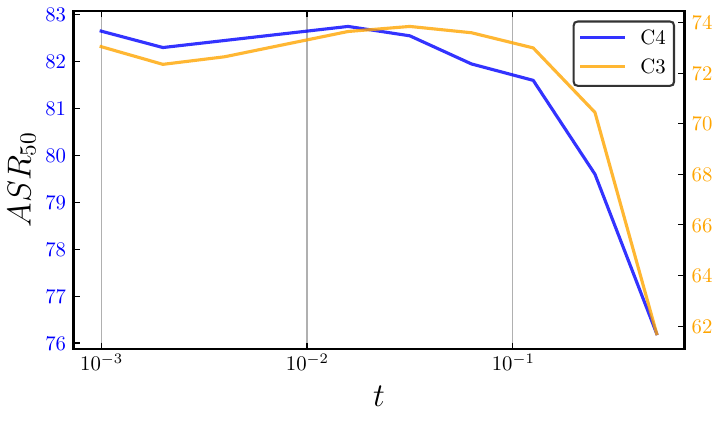}
    \includegraphics[width=0.45\textwidth]{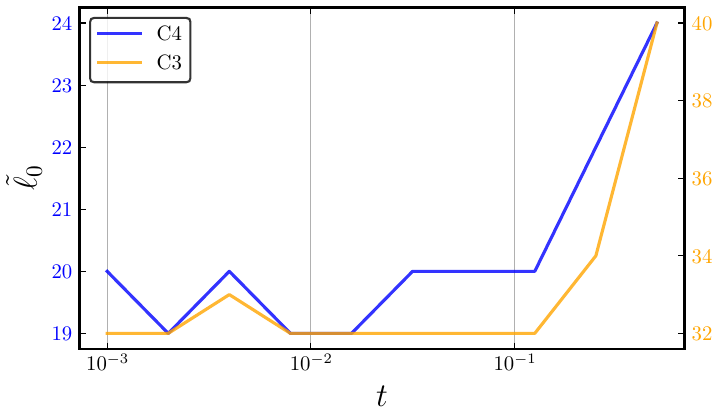}
    \caption{Ablation study on  $t$ for CIFAR-10 \croceloneshort and \jiangloneshort. For each, we report the attack success rate at the 50 feature budget (\textit{left}) and the median \lzero norm of the adversarial perturbation (\textit{right}).}
    \label{fig:ablation_tau_plus}
\end{figure*}

We also conducted an ablation study on the sparsity threshold adjustment factor $t$ used to adaptively update $\tau$. 
In the following we keep the default values for $\tau_0=0.3$ and $\sigma=10^{-3}$. 
We executed the attack on $1000$ randomly selected samples against \croceloneshort and \jiangloneshort models and recorded the $ASR_{50}$ and the median $l_0$ norm of the resulting adversarial perturbations. 
In \autoref{fig:ablation_tau_plus}, we once against observe the robustness of \sigmazero to this parameter, yielding similar and effective results when $t \leq 10^{-1}$.

Overall, the ablation study revealed consistent trends across the distinct models and datasets. 
In all cases, we identified a broad parameter configuration range where our attack maintained robustness to the hyperparameter selection, making hyperparameter optimization for the attacker a swift task. 
This robustness is further evidenced by the results presented in the experimental comparisons, where our attack consistently outperforms competing attacks even with a shared hyperparameter configuration across all models.

\section{Additional Experimental Comparisons}\label{appendix:additional_experiments}
\subsection{Comparisons with Minimum-norm Attack}\label{appendix:hundred_steps}
In our experimental setup, we also consider a reduced number of queries, to test whether the attack can also run faster while remaining effective. We thus replicate our experimental comparison involving \sigmazero and state-of-the-art sparse attacks while restricting the number of steps to $N=100$. 
The results are summarized in Tables~\ref{tab:0.1k_minimum_norm_mnist}-\ref{tab:0.1k_minimum_norm_imagenet}. 
Compared to the results with $N=1000$ steps reported in Tables~\ref{tab:1k_minimum_norm_complete} and~\ref{tab:1k_remaining_minimum_norm_cifar_imagenet}, the ASR of most competitive attacks decreases, while \sigmazero remains effective by consistently reaching an ASR of 100\%.
This shows that \sigmazero remains an effective, reliable and fast approach to crafting minimum-norm attacks even with reduced query budgets.


\subsection{Comparisons with Fixed-budget Attacks}\label{appendix:maxattacks}
Fixed-budget attacks, i.e., Sparse-RS~\citep{Croce22SparseRS}, \PGDlzero\citep{Croce2019SparseAI}, and Sparse-PGD~\citep{zhong2024towards}, have been designed to generate sparse adversarial perturbations given a fixed-budget $k$, therefore, drawing comparisons with minimum-norm attacks is not a straightforward task. 
Specifically, in their threat model, the attacker imposes a maximum limit on the number of perturbed features, and the attack then outputs the adversarial example that minimizes the model's confidence in predicting the true label of the sample.
However, since the fixed-budget threat model differs from the minimum-norm scenario we consider in this paper, which does not assume a maximum norm value $k$, we evaluate \sigmazero in a fixed-budget fashion by discarding all adversarial perturbations that exceed $k$. 
Furthermore, as for fixed-budget attack, we let \sigmazero to leverage the input parameter $k$ to early stop the optimization procedure and reduce the number of consumed queries to the target model. 
Throughout this evaluation, the number of steps taken by Sparse-RS is always doubled compared to the other two white-box attacks, as it does not utilize the backward pass employed by the others. 
The main paper reports to this end an evaluation of \sigmazero in a fixed-budget approach (cf. Tables~\ref{tab:cifar-fixed_sigma-k}-\ref{tab:imagenet-fixed_sigma-k}).
The remaining experiments, involving additional models for CIFAR-10 and ImageNet, are reported in Tables~\ref{tab:1k_remaining_maximum_budget_cifar}-\ref{tab:1k_remaining_maximum_budget_imagenet}. 
Furthermore, to explore the effects of increased iterations on convergence and success rate, we increased the number of iterations up to $N=10000$ (Tables~\ref{tab:5k_maximum_budget_mnist}-\ref{tab:10k_steps_cifar_imgnet}), while always doubling the iterations for Sparse-RS. 
These additional experiments cover the three datasets MNIST, CIFAR-10, and ImageNet, 18 distinct models, and various feature budgets.
The results again affirm that, \sigmazero consistently outperforms competing approaches or synergizes well with them for a comprehensive robustness assessment.

\subsection{Robustness Evaluation Curves}\label{appendix:sec_curves}
\begin{figure}[h!]
    \centering
    \includegraphics[width=0.6\textwidth]{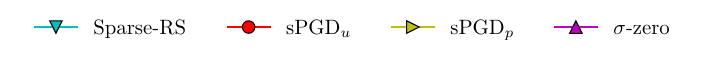}\vspace{-0.5em}
    \includegraphics[width=0.45\textwidth]{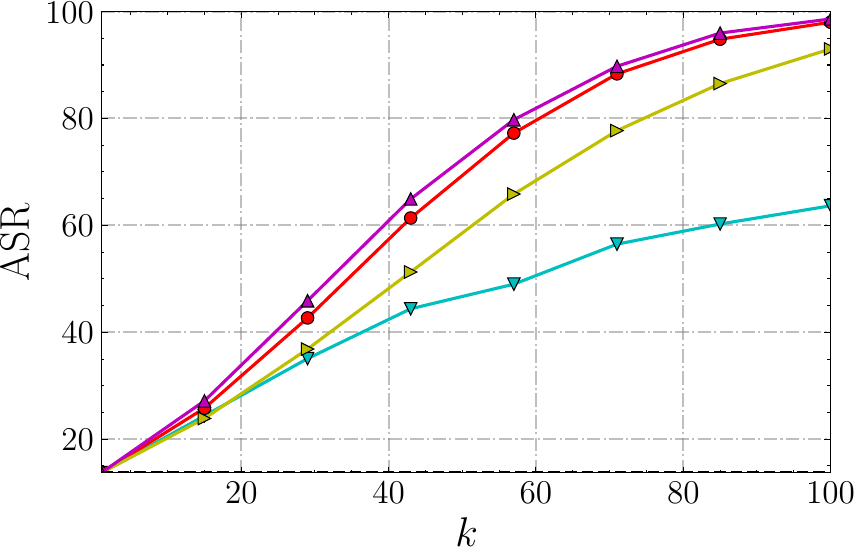}\hfill
    \includegraphics[width=0.45\textwidth]{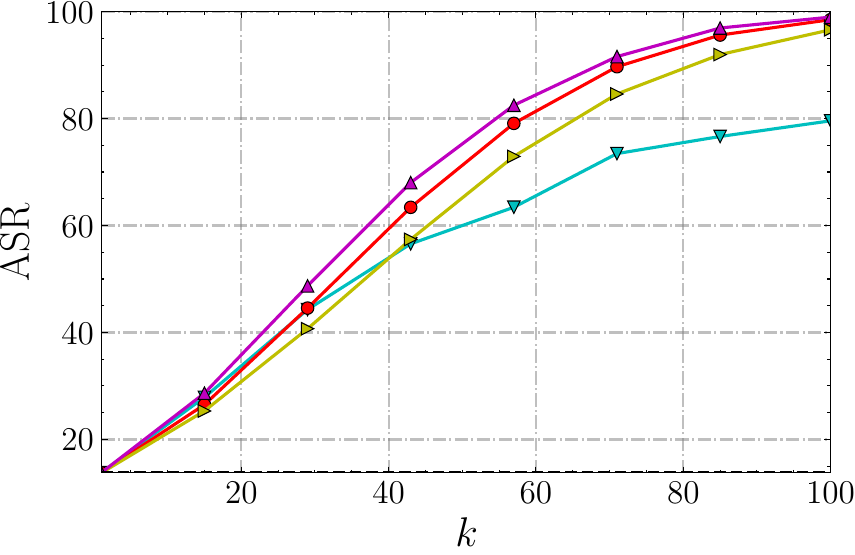}
    \caption{Robustness evaluation curves for fixed-budget attacks on \croceloneshort. For each budget level $k$, each attack has been run with $1000$ iterations (left-most plot) and $5000$ iterations (right-most plot). Sparse-RS has been run with double the iterations as it relies solely on forward calls.}
    \label{fig:sec_curves_appendix_fixed}
\end{figure}

\begin{figure}[h!]
    \centering
    \includegraphics[width=1\textwidth]{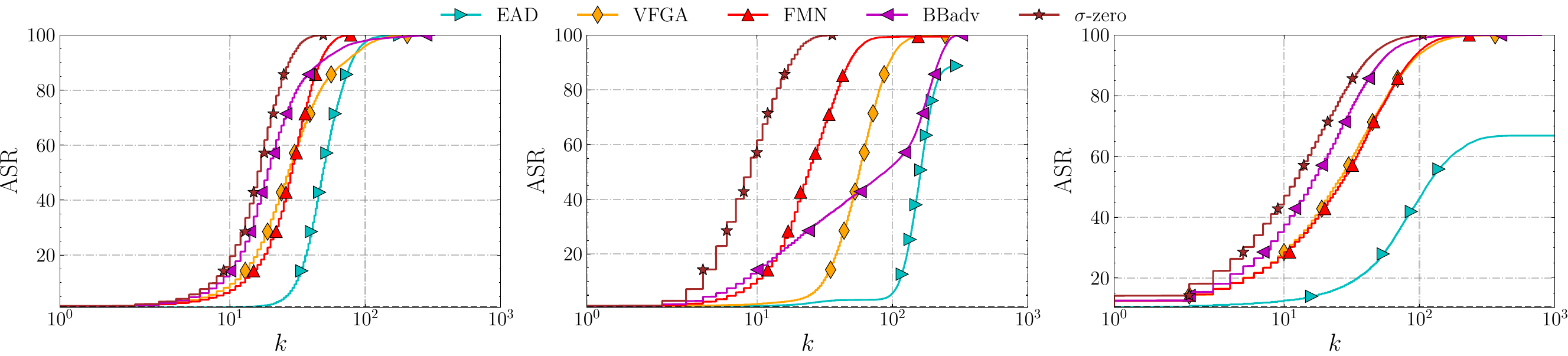}\hfill
    \includegraphics[width=1\textwidth]{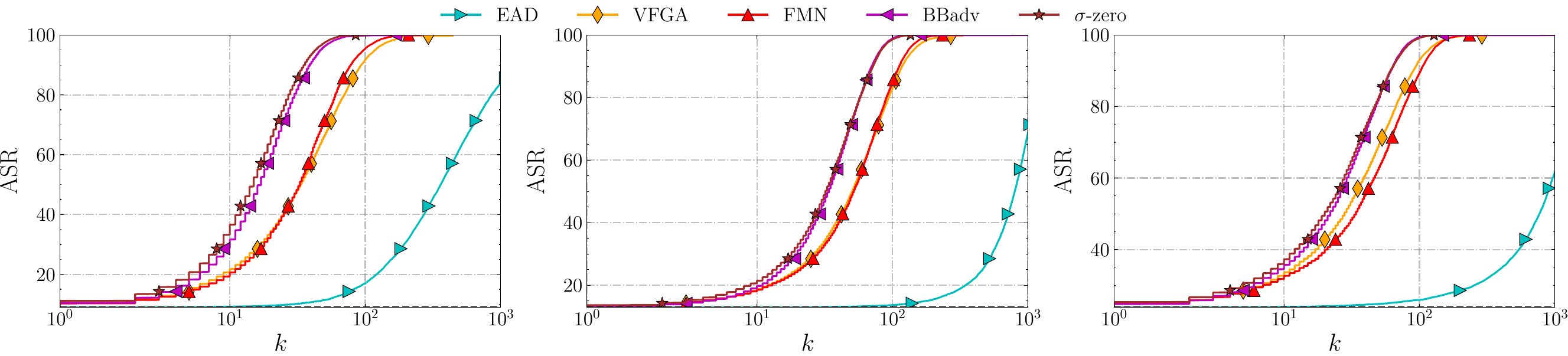}\hfill
    \includegraphics[width=1\textwidth]{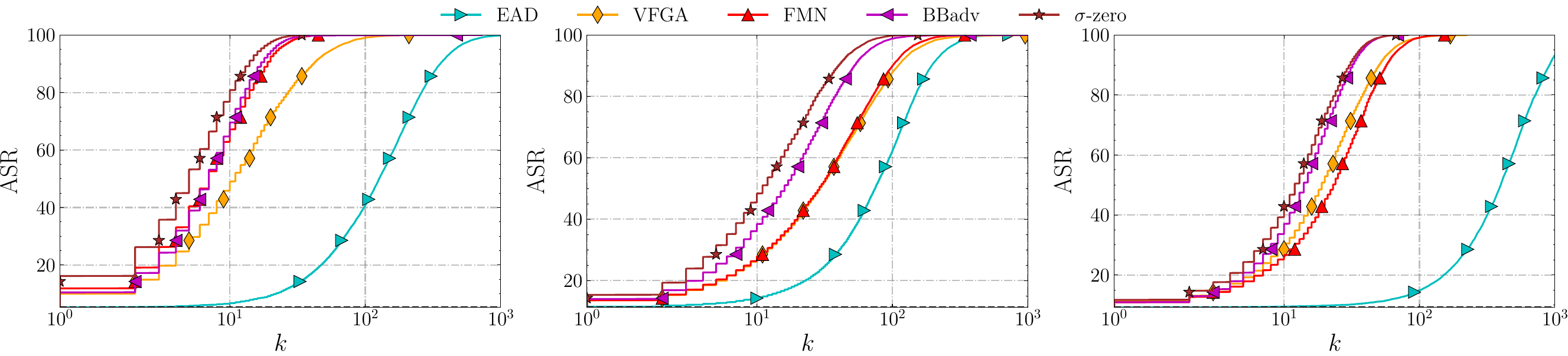}\hfill
    
    \caption{
    From the leftmost to the rightmost we report the robustness evaluation curves for \smallcnnddnshort, \smallcnntradesshort, \carmonshort (top-row), \augustinshort, \croceloneshort, \jiangloneshort (middle-row) and \standardshort, \gowalshort, \engstromshort (bottom-row).}
    \label{fig:sec_curves_appendix_1}
\end{figure}

\begin{figure}[h!]
    \centering
    \includegraphics[width=1\textwidth]{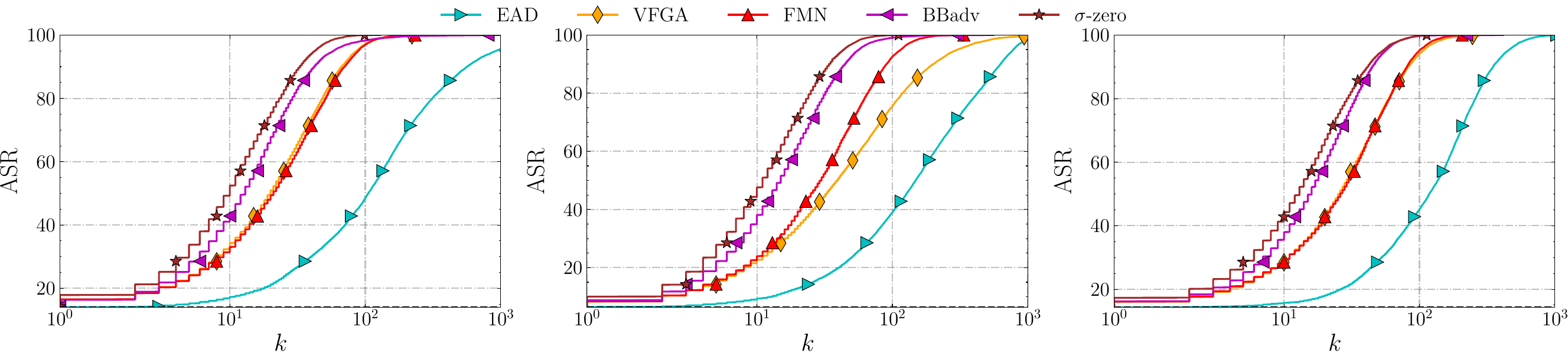}\hfill
    \includegraphics[width=1\textwidth]{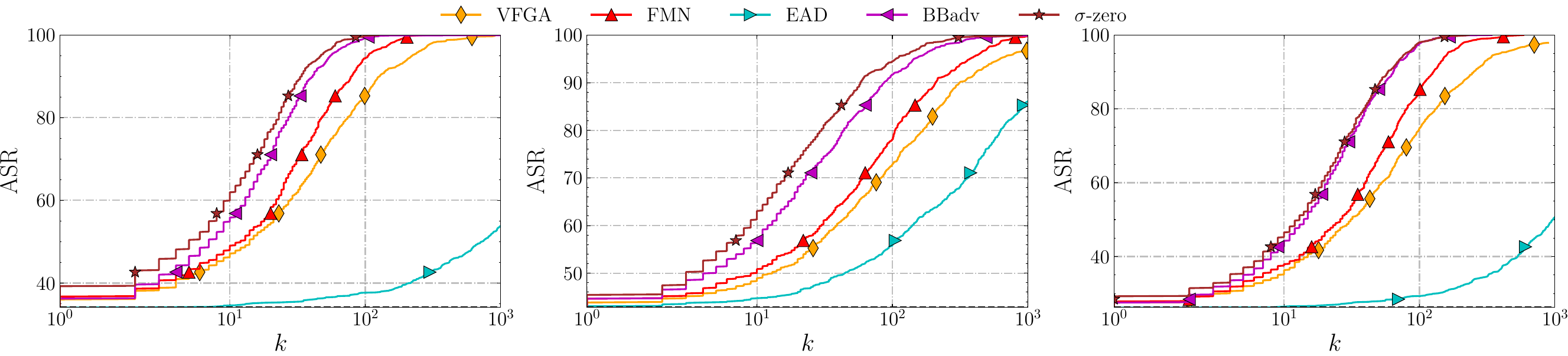}\hfill
    \includegraphics[width=1\textwidth]{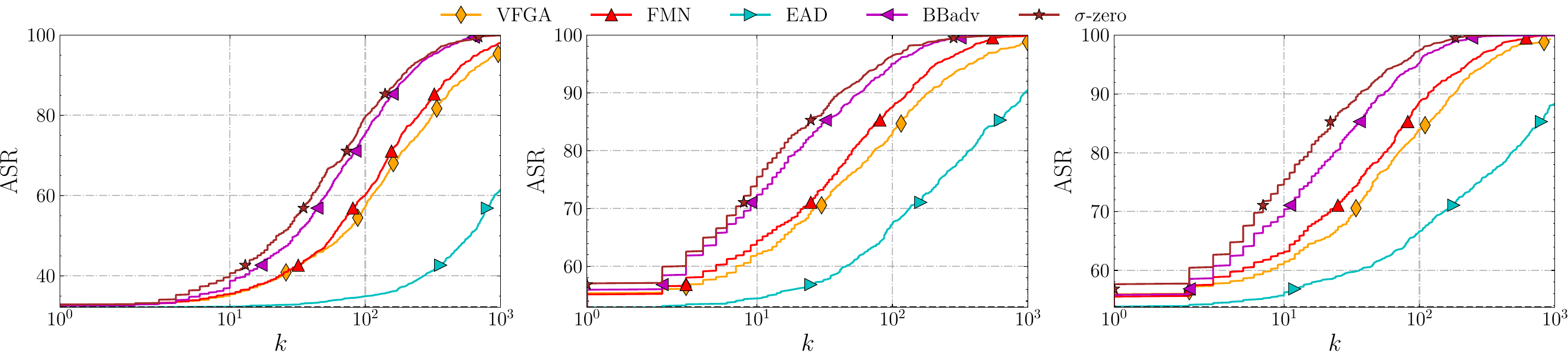}\hfill
    
    \caption{From the leftmost to the rightmost we report the robustness evaluation curves for \chenshort, \xushort, \addepallishort (top-row), \resnetvulnerableshort, \engstromimagenetshort, \hendrycksshort (middle-row) and \debenedettishort, \wongshort, \salmanshort (bottom-row)}
    \label{fig:sec_curves_appendix_2}
\end{figure}

\rebuttal{We provide robustness evaluation curves for fixed-budget attacks on a CIFAR-10 model (\croceloneshort), running each attack multiple times across various perturbation budgets $k$. The number of iterations is set to $N=1000$ and $N=5000$, with Sparse-RS allocating twice the iterations due to its reliance solely on forward passes.
The results, depicted in \autoref{fig:sec_curves_appendix_fixed}, demonstrates that \sigmazero consistently outperforms fixed-budget attacks across all perturbation budgets $k$.}
Additionally, we present in Figs.~\ref{fig:sec_curves_appendix_1}-\ref{fig:sec_curves_appendix_2} the robustness evaluation curves depicting the performance of minimum-norm \lzero-attacks against all the models analyzed in our paper. These findings reinforce our experimental analysis, explicitly demonstrating that the \sigmazero attack consistently achieves higher values of ASR while employing smaller \lzero-norm perturbations compared to alternative attacks.

\begin{table}[htb]
\caption{Minimum-norm comparison results for CIFAR-10 and ImageNet with $N=1000$ on remaining models. 
For each attack and model ({M}), we report ASR at $k=10, 50, \infty$, median perturbation size \medianlzero, mean runtime $s$ (in seconds), mean number of queries $q$ ($\div$ 1000), and maximum VRAM usage (in GB). When VFGA exceeds the VRAM limit, we re-run it using a smaller batch size, increasing its runtime $t$. We denote those cases with the symbol `$\star$'. Lastly we indicate with \sigmazero$^*$ the case where we use $\sigma = 1$ and $\tau_0 = 0.1$.}
\label{tab:1k_remaining_minimum_norm_cifar_imagenet}
\resizebox{\textwidth}{!}{
\renewcommand{\arraystretch}{1.1}
  \setlength\tabcolsep{1.5pt} 

\begin{tabular}{c|c|cccc|ccc|c|cccc|ccc}
\toprule
\multicolumn{1}{c}{\textbf{Attack}} & \multicolumn{1}{c}{\textbf{M}}& \multicolumn{1}{c}{\textbf{ASR$_{24}$}} & \multicolumn{1}{c}{\textbf{ASR$_{50}$}} & \multicolumn{1}{c}{\textbf{ASR$_\infty$}} & \multicolumn{1}{c}{\medianlzero} & \multicolumn{1}{c}{$\mathbf{s}$} & \multicolumn{1}{c}{$\mathbf{q}$}  & \multicolumn{1}{c}{\textbf{VRAM}} & \multicolumn{1}{c}{\textbf{M}} & \multicolumn{1}{c}{\textbf{ASR$_{24}$}} & \multicolumn{1}{c}{\textbf{ASR$_{50}$}} & \multicolumn{1}{c}{\textbf{ASR$_\infty$}} & \multicolumn{1}{c}{\medianlzero} & \multicolumn{1}{c}{$\mathbf{s}$} & \multicolumn{1}{c}{$\mathbf{q}$}  & \multicolumn{1}{c}{\textbf{VRAM}}\\ \hline\hline
\multicolumn{17}{c}{CIFAR-10} \\ \hline\hline
SF & \multirow{8}{*}{\standardshort} & 11.19 & 11.19 & 56.56 & 3072 & 1.42 & 0.37 & 1.57 & \multirow{8}{*}{\chenshort} & 23.67 & 24.85 & 62.41 & 3072 & 9.86 & 0.20 & 5.50 \\
EAD &  & 10.91 & 21.33 & 100.0 & 126 & 2.32 & 6.90 & 1.47 &  & 23.23 & 33.68 & 100.0 & 105 & 8.33 & 5.37 & 5.39 \\
PDPGD &  & 41.70 & 78.97 & 100.0 & 27 & 0.64 & 2.00 & 1.31 &  & 33.38 & 48.96 & 99.82 & 51 & 2.15 & 2.00 & 5.12 \\
VFGA &  & 77.25 & 93.41 & 99.99 & 11 & 0.17 & 0.32 & 11.96 &  & 56.76 & 81.79 & 99.89 & 20 & 4.30$^\star$ & 0.62 & \minfty \\
FMN &  & 95.99 & 99.97 & 100.0 & 7 & 0.60 & 2.00 & 1.3 &  & 54.74 & 79.70 & 100.0 & 21 & 2.05 & 2.00 & 5.12 \\
BB &  & 97.43 & 99.79 & 100.0 & 7 & 5.81 & 2.76 & 1.47 &  & 59.82 & 78.76 & 83.58 & 16 & 12.49 & 3.14 & 5.39 \\
BBadv &  & 97.50 & 99.86 & 100.0 & 7 & 4.57 & 2.01 & 1.63 &  & 74.51 & 93.42 & 100.0 & 13 & 6.99 & 2.01 & 5.51 \\
\cdashline{1-17} 
\rc\sigmazero &  & \textbf{99.20} & \textbf{100.0} & 100.0 & 5 & 0.74 & 2.00 & 1.51 &  & \textbf{81.23} & \textbf{97.33} & 100.0 & \textbf{10} & 2.75 & 2.00 & 5.90 \\ \hline

SF & \multirow{8}{*}{\gowalshort} & 16.78 & 16.79 & 35.38 & \binfty & 19.74 & 0.62 & 10.00 & \multirow{8}{*}{\xushort} & 12.12 & 12.14 & 70.77 & 3072 & 3.28 & 0.22 & 2.25 \\
EAD &  & 20.75 & 35.90 & 100.0 & 74 & 10.76 & 5.55 & 9.92 &  & 14.51 & 23.62 & 100.0 & 148 & 2.23 & 5.80 & 2.15 \\
PDPGD &  & 23.84 & 40.89 & 100.0 & 69 & 3.96 & 2.00 & 8.86 &  & 25.31 & 38.41 & 100.0 & 69 & 0.76 & 2.00 & 2.0 \\
VFGA &  & 45.28 & 67.51 & 99.88 & 29 & 4.91$^\star$ & 1.02 & \minfty &  & 38.42 & 56.72 & 99.81 & 39 & 3.15$^\star$ & 1.84 & \minfty \\
FMN &  & 45.73 & 68.38 & 100.0 & 29 & 3.91 & 2.00 & 8.86 &  & 44.38 & 70.24 & 100.0 & 30 & 0.73 & 2.00 & 2.0 \\
BB &  & 15.26 & 17.14 & 17.94 & \binfty & 3.46 & 2.08 & 9.93 &  & 70.11 & 93.24 & 100.0 & 15 & 6.49 & 2.87 & 2.16 \\
BBadv &  & 64.47 & 88.92 & 100.0 & 16 & 8.85 & 2.01 & 10.03 &  & 69.45 & 92.91 & 100.0 & 15 & 6.02 & 2.01 & 2.22 \\
\cdashline{1-17}
\rc\sigmazero &  & \textbf{75.63} & \textbf{94.47} & 100.0 & \textbf{11} & 4.41 & 2.00 & 10.43 &  & \textbf{79.59} & \textbf{96.93} & 100.0 & \textbf{11} & 0.89 & 2.00 & 2.65 \\ \hline
{SF} &  & 29.51 & 40.86 & 93.82 & 3039 & 9.3 & 1.56 & 1.90 &  & 25.88 & 26.54 & 51.80 & 3072 & 0.58 & 0.33 & 0.51 \\
{EAD} &  & 9.92 & 11.14 & 100.0 & 398 & 2.57 & 5.66 & 1.89 &  & 19.44 & 29.23 & 100.0 & 118 & 1.01 & 5.32 & 0.41 \\
{PDPGD} &  & 32.60 & 49.19 & 100.0 & 51 & 1.16 & 2.00 & 1.8 &  & 29.98 & 41.00 & 100.0 & 66 & 0.44 & 2.00 & 0.36 \\
{VFGA} &  & 61.19 & 90.04 & 99.88 & 19 & 0.28 & 0.52 & 16.53 &  & 48.63 & 74.15 & 99.54 & 25 & 0.17 & 0.77 & 3.07 \\
{FMN} &  & 52.14 & 85.60 & 100.0 & 23 & 1.09 & 2.00 & 1.8 &  & 47.89 & 73.71 & 100.0 & 26 & 0.41 & 2.00 & 0.36 \\
{BB} &  & 21.44 & 31.03 & 31.36 & \binfty & 3.01 & 2.37 & 1.89 &  & 68.37 & 91.83 & 100.0 & 15 & 10.90 & 2.93 & 0.41 \\
{BBadv} &  & 77.88 & 99.11 & 100.0 & 14 & 4.51 & 2.01 & 1.99 &  & 67.35 & 93.04 & 100.0 & 16 & 4.60 & 2.01 & 0.54 \\
\cdashline{1-17}
\rc\sigmazero & \multirow{-8}{*}{\textit \engstromshort} & \textbf{81.38} & \textbf{99.15} & 100.0 & \textbf{12} & 1.39 & 2.00 & 1.91 & \multirow{-8}{*}{\textit \addepallishort} & \textbf{73.96} & \textbf{94.21} & 100.0 & \textbf{13} & 0.63 & 2.00 & 0.51 \\ \hline
FMN & \multirow{4}{*}{\zhongtradesshort} & 39.57 & 74.75 & 100.0 & 32 & 0.11 & 2.00 & 0.59 & \multirow{4}{*}{\zhongsatshort} & 48.3 & 78.16 & 100.0 & 26 & 0.11 & 2.00 & 0.59 \\
BBadv &  & 14.07 & 18.57 & 100.0 & 183 & 2.52 & 2.01 & 0.65 &  & 18.33 & 19.75 & 100.0 & 290 & 2.57 & 2.01 & 0.65 \\

\sigmazero &  & 12.38 & 15.91 & 100.0 & 144 & 0.24 & 2.00 & 1.03 &  & 18.52 & 21.31 & 100.0 & 187 & 0.33 & 2.00 & 1.03 \\
\cdashline{1-17}\rc\sigmazero$^*$ &  & \textbf{44.78} & \textbf{85.05} & 100.0 & \textbf{27} & 0.23 & 2.00 & 1.03 &  & \textbf{54.62} & \textbf{90.12} & 100.0 & \textbf{22} & 0.23 & 2.00 & 1.03 
\\
\hline\hline
\multicolumn{17}{c}{ImageNet} \\ \hline\hline
\multicolumn{1}{c}{\textbf{Attack}} & \multicolumn{1}{c}{\textbf{M}}& \multicolumn{1}{c}{\textbf{ASR$_{24}$}} & \multicolumn{1}{c}{\textbf{ASR$_{50}$}} & \multicolumn{1}{c}{\textbf{ASR$_\infty$}} & \multicolumn{1}{c}{\medianlzero} & \multicolumn{1}{c}{$\mathbf{s}$} & \multicolumn{1}{c}{$\mathbf{q}$}  & \multicolumn{1}{c}{\textbf{VRAM}} & \multicolumn{1}{c}{\textbf{M}} & \multicolumn{1}{c}{\textbf{ASR$_{24}$}} & \multicolumn{1}{c}{\textbf{ASR$_{50}$}} & \multicolumn{1}{c}{\textbf{ASR$_\infty$}} & \multicolumn{1}{c}{\medianlzero} & \multicolumn{1}{c}{$\mathbf{s}$} & \multicolumn{1}{c}{$\mathbf{q}$}  & \multicolumn{1}{c}{\textbf{VRAM}}\\ \hline

EAD &  & 56.6 & 60.2 & 100.0 & 0 & 21.38 & 5.50 & 1.41 &  & 59.0 & 61.4 & 100.0 & 0 & 7.89 & 5.29 & 0.48 \\
VFGA &  & 69.0 & 76.2 & 98.8 & 0 & 6.11$^\star$ & 1.43 & \minfty &  & 66.8 & 76.6 & 99.3 & 0 & 1.74$^\star$ & 1.21 & \minfty \\
FMN &  & 71.0 & 79.5 & 100.0 & 0 & 1.97 & 2.00 & 2.30 &  & 70.9 & 78.7 & 100.0 & 0 & 0.72 & 2.00 & 0.67 \\
BBadv &  & 82.3 & 89.0 & 100.0 & 0 & 185.34 & 2.01 & 2.41 &  & 80.3 & 89.6 & 100 & 0 & 199.47 & 2.01 & 0.73 \\
\cdashline{1-17} 
\rc\sigmazero & \multirow{-5}{*}{\wongshort} & \textbf{85.1} & \textbf{91.4} & 100.0 & 0 & 2.76 & 2.00 & 2.52 & \multirow{-5}{*}{\salmanshort} & \textbf{86.2} & \textbf{92.8} & 100.0 & 0 & 1.13 & 2.00 & 0.84 \\ \hline
FMN & \multirow{3}{*}{\pengshort} & 39.2 & 48.5 & 100 & 54.5 & 5.84 & 2 & 17.44 & \multirow{3}{*}{\moshort}  & 38.1 & 46.8 & 100 & 67 & 5.17 & 2 & 7.91 \\
BBadv &  & 49.7 & 62 & 100 & 25.5 & 128.2 & 2 & 17.86 &  & 46.5 & 58.6 & 100 & 29.5 & 113.87 & 2 & 8.30 \\
\cdashline{1-17}\rc
\sigmazero &  & \textbf{55.4} & \textbf{68.2} & 100 & \textbf{16} & 8.62 & 2 & 19.03 & & \textbf{50.1} & \textbf{64.4} & 100 & \textbf{24} & 6.6 & 2 & 9.47 \\
\bottomrule
\end{tabular}
}
\end{table}

\begin{table*}[htbp]
\caption{Minimum-norm comparison results for MNIST with $N=100$. See the caption of \autoref{tab:1k_remaining_minimum_norm_cifar_imagenet} for further details.}
\label{tab:0.1k_minimum_norm_mnist}
\centering
\resizebox{0.99\textwidth}{!}{%
  \setlength\tabcolsep{1.5pt} 

\begin{tabular}{r|c|cccc|ccc|c|cccc|ccc}
\toprule
\multicolumn{1}{c}{\textbf{Attack}} & \multicolumn{1}{c}{\textbf{M}}& \multicolumn{1}{c}{\textbf{ASR$_{24}$}} & \multicolumn{1}{c}{\textbf{ASR$_{50}$}} & \multicolumn{1}{c}{\textbf{ASR$_\infty$}} & \multicolumn{1}{c}{\medianlzero} & \multicolumn{1}{c}{$\mathbf{s}$} & \multicolumn{1}{c}{$\mathbf{q}$}  & \multicolumn{1}{c}{\textbf{VRAM}} & \multicolumn{1}{c}{\textbf{M}} & \multicolumn{1}{c}{\textbf{ASR$_{24}$}} & \multicolumn{1}{c}{\textbf{ASR$_{50}$}} & \multicolumn{1}{c}{\textbf{ASR$_\infty$}} & \multicolumn{1}{c}{\medianlzero} & \multicolumn{1}{c}{$\mathbf{s}$} & \multicolumn{1}{c}{$\mathbf{q}$}  & \multicolumn{1}{c}{\textbf{VRAM}}\\
\hline\hline
\multicolumn{17}{c}{MNIST} \\ \hline\hline
{SF} &  & 5.11 & 6.76 & 96.98 & 469 & 1.07 & 0.18 & 0.07 &  & 0.98 & 1.21 & 91.68 & 463 & 2.87 & 0.86 & 0.07 \\
EAD  &  & 3.73 & 46.65 & 100.0 & 52 & 0.06 & 1.14 & 0.07 &  & 3.51 & 35.57 & 100.0& 61 & 0.06 & 0.99 & 0.07 \\
{PDPGD} &  & 0.98 & 0.98 & 100.0 & 359 & 0.01 & 0.20 & 0.07 &  & 0.52 & 0.52 & 95.02 & 254 & 0.01 & 0.20 & 0.07 \\
{VFGA} &  & 4.82 & 82.68 & 100.0 & 27 & 0.07 & 0.76 & 0.23 &  & 4.82 & 38.99 & 99,98 & 57 & 0.07 & 1.34 & 0.24 \\
{BB} &  & \textbf{68.52} & 98.00 & 100.0 & \textbf{20} & 0.13 & 1.19 & 0.08 &  & 62.98 & 83.00 & 87.87 & 18 & 0.13 & 1.69 & 0.08 \\
{FMN} &  & 33.03 & 83.09 & 88.92 & 30 & 0.01 & 0.20 & 0.07 &  & 10.05 & 14.03 & 14.81 & \binfty & 0.01 & 0.20 & 0.07 \\
{BBadv} &  & 62.29 & 90.88 & 100.0 & 21 & 0.09 & 0.21 & 0.08 &  & 41.19 & 58.80 & 100.0& 34 & 0.07 & 0.21 & 0.08 \\
\cdashline{1-17}
\rc\sigmazero & \multirow{-8}{*}{\textit \smallcnnddnshort} & 61.12 & \textbf{98.45} & 100.0 & 22 & 0.01 & 0.20 & 0.08 & \multirow{-8}{*}{\textit \smallcnntradesshort} & \textbf{87.20} & \textbf{99.82} & 100.0& \textbf{13} & 0.01 & 0.20 & 0.08 \\ \bottomrule
\end{tabular}
}
\end{table*}

\begin{table*}[!htb]
\caption{Minimum-norm comparison results for CIFAR-10 with $N=100$. See the caption of \autoref{tab:1k_remaining_minimum_norm_cifar_imagenet} for further details.}
\label{tab:exp_results_cifar_appendix}
\centering
\resizebox{0.99\textwidth}{!}{%
\renewcommand{\arraystretch}{1}
  \setlength\tabcolsep{1.5pt} 

\begin{tabular}{r|c|cccc|ccc|c|cccc|ccc}
\toprule
\multicolumn{1}{c}{\textbf{Attack}} & \multicolumn{1}{c}{\textbf{M}}& \multicolumn{1}{c}{\textbf{ASR$_{24}$}} & \multicolumn{1}{c}{\textbf{ASR$_{50}$}} & \multicolumn{1}{c}{\textbf{ASR$_\infty$}} & \multicolumn{1}{c}{\medianlzero} & \multicolumn{1}{c}{$\mathbf{s}$} & \multicolumn{1}{c}{$\mathbf{q}$}  & \multicolumn{1}{c}{\textbf{VRAM}} & \multicolumn{1}{c}{\textbf{M}} & \multicolumn{1}{c}{\textbf{ASR$_{24}$}} & \multicolumn{1}{c}{\textbf{ASR$_{50}$}} & \multicolumn{1}{c}{\textbf{ASR$_\infty$}} & \multicolumn{1}{c}{\medianlzero} & \multicolumn{1}{c}{$\mathbf{s}$} & \multicolumn{1}{c}{$\mathbf{q}$}  & \multicolumn{1}{c}{\textbf{VRAM}}\\
\hline\hline
\multicolumn{17}{c}{CIFAR-10} \\ \hline\hline

{SF} &  & 17.67 & 17.76 & 47.26 & \binfty & 3.17 & 0.35 & 1.62 &  & 16.75 & 16.79 & 35.36 & \binfty & 19.74 & 0.62 & 10.00 \\ %
{EAD} &  & 16.74 & 28.74 & 100.0 & 100.0& 0.27 & 0.80 & 1.53  &  & 19.79 & 32.94 & 100.0& 83 & 1.58 & 0.82 & 10.04 \\%
{PDPGD} &  & 10.31 & 10.31 & 99.39 & 2421 & 0.05 & 0.20 & 1.43 &  & 11.26 & 11.26 & 99.75 & 2814 & 0.32 & 0.2 & 8.97 \\ %
{VFGA} &  & 50.73 & 75.34 & 93.69 & 24 & 0.23 & 0.72 & 11.83 &  & 45.33 & 67.05 & 87.75 & 29 & 3.75 & 0.86 & \minfty \\ %
{FMN} &  & 46.90 & 69.36 & 80.68 & 27 & 0.05 & 0.20 & 1.43  &  & 42.67 & 61.49 & 72.34 & 33 & 0.31 & 0.2 & 8.98 \\%
{BB} &  & 12.98 & 14.29 & 14.97 & \binfty & 0.44 & 1.95 & 1.59  &  & 14.99 & 16,88 & 17.91 & \binfty & 2.67 & 1.95 & 10.04 \\%
{BBadv} &  & 60.52 & 86.63 & 100.0 & 18 & 0.41 & 0.21 & 1.59 &  & 59.61 & 84.59 & 100.0& 18 & 0.76 & 0.21 & 10.04 \\%
\cdashline{1-17}
\rc\sigmazero & \multirow{-8}{*}{\textit \carmonshort} & \textbf{63.60} & \textbf{88.27} & 100.0 & \textbf{16} & 0.08 & 0.20 & 1.84 & \multirow{-8}{*}{\textit \gowalshort} & \textbf{63.44} & \textbf{87.2} & 100.0& \textbf{16} & 0.44 & 0.2 & 10.29 \\ \hline%

{SF} &  & 17.86 & 20.59 & 94.26 & 3071 & 2.44 & 0.26 & 1.91 &  & 21.07 & 38.76 & 82.71 & 3062 & 4.30 & 9.67 & 1.90\\%
{EAD} &  & 9.50 & 10.67 & 100.0 & 451 & 0.30 & 0.71 & 2.01 &  & 9.68 & 10.56 & 100.0 & 434 & 0.48 & 0.90 & 2.00 \\%
{PDPGD} &  & 8.92 & 8.92 & 75.31 & 3052 & 0.09 & 0.20 & 1.91 &  & 9.17 & 9.17 & 99.90 & 2709 & 0.12 & 0.20 & 1.91 \\%
{VFGA} &  & 39.31 & 66.46 & 91.64 & 33 & 0.34 & 0.87 & 16.64 &  & 60.94 & 90.04 & 99.16 & 19 & 0.29 & 0.52 & 16.64\\ %
{FMN} &  & 37.13 & 62.41 & 71.3 & 36 & 0.08 & 0.20 & 1.92  &  & 50.70 & 79.48 & 87.20 & 24 & 0.08 & 0.20 & 1.91\\ %
{BB} &  & 38.18 & 53.53 & 57.05 & 40 & 0.59 & 1.90 & 2.00 &  & 26.39 & 32.41 & 32.83 & \binfty & 0.50 & 1.93 & 2.00 \\%
{BBadv} &  & \textbf{63.56} & \textbf{92.74} & 100.0 & \textbf{19} & 0.40 & 0.21 & 2.00 &  & \textbf{74.91} & \textbf{98.37} & 100.0 & \textbf{15} & 0.41 & 0.21 & 2.00  \\%
\cdashline{1-17}
\rc\sigmazero & \multirow{-8}{*}{\textit \augustinshort} & 56.94 & 88.60 & 100.0 & 21 & 0.11 & 0.20 & 2.25 & \multirow{-8}{*}{\textit \engstromshort} & 68.14 & 94.90 & 100.0 & 16 & 0.11 & 0.20 & 2.25 \\ \hline

{SF} &  & 20.89 & 24.36 & 58.29 & 3072 & 1.63 & 0.48 & 0.66 &  & 23.87 & 24.85 & 62.42 & 3072 & 9.86 & 0.2 & 5.50 \\
{EAD} &  & 13.03 & 13.18 & 100.0 & 835 & 0.11 & 0.65 & 0.64&  & 21.71 & 29.59 & 100.0& 128 & 0.67 & 0.66 & 5.51 \\
{PDPGD} &  & 12.95 & 12.98 & 99.47 & 2566 & 0.04 & 0.20 & 0.59&  & 13.96 & 13.96 & 54.16 & 3072 & 0.21 & 0.2 & 5.23 \\
{VFGA} &  & 28.63 & 49.73 & 82.94 & 51 & 0.13 & 1.13 & 4.44 &  & 56.81 & 82.04 & 97.08 & 20 & 4.32 & 0.61 & \minfty \\
{FMN} &  & 26.76 & 37.90 & 43.90 & \binfty & 0.03 & 0.20 & 0.59&  & 53.42 & 76.59 & 87.1 & 22.0 & 0.21 & 0.2 & 5.24 \\
{BB} &  & 16.40 & 22.91 & 27.64 & \binfty & 1.04 & 2.25 & 0.65&  & 60.74 & 78.14 & 84.46 & 17 & 1.36 & 1.67 & 5.55 \\
{BBadv} &  & \textbf{33.68} & \textbf{66.79} & 100.0 & \textbf{37} & 0.40 & 0.21 & 0.65 &  & \textbf{70.31} & 91.58 & 100.0& 14 & 0.55 & 0.21 & 5.51 \\
\cdashline{1-17}
\rc\sigmazero & \multirow{-8}{*}{\textit \croceloneshort} & 30.56 & 57.71 & 100.0 & 43 & 0.04 & 0.20 & 0.89 & \multirow{-8}{*}{\textit \chenshort} & 69.49 & \textbf{91.87} & 100.0& \textbf{14} & 0.25 & 0.2 & 6.76 \\ \hline

{SF} &  & 31.85 & 42.97 & 84.45 & 70 & 1.54 & 0.47 & 0.66 &  & 12.03 & 12.14 & 70.77 & 3072 & 3.28 & 0.22 & 2.25 \\
{EAD}  &  & 24.1 & 24.4 & 100.0& 844 & 0.12 & 0.66 & 0.65  &  & 13.61 & 21.61 & 100.0& 162 & 0.31 & 0.8 & 2.27 \\
{PDPGD}  &  & 23.78 & 23.78 & 66.62 & 3072 & 0.04 & 0.2 & 0.59  &  & 6.31 & 6.31 & 96.2 & 2773 & 0.06 & 0.2 & 2.11 \\
{VFGA}  &  & 46.7 & 69.52 & 93.05 & 28 & 0.14 & 0.77 & 4.22 &  & 38.22 & 56.56 & 75.79 & 39.5 & 1.45 & 1.06 & \minfty \\
{FMN} &  & 42.69 & 58.78 & 65.83 & 35 & 0.03 & 0.2 & 0.59 &  & 40.27 & 59.69 & 68.88 & 35 & 0.06 & 0.2 & 2.19 \\
{BB} &  & 25.91 & 27.98 & 29.51 & \binfty & 0.54 & 2.09 & 0.65 &  & 66.02 & 90.74 & 100.0& 16 & 0.65 & 1.07 & 2.27 \\
{BBadv} &  & \textbf{52.25} & \textbf{80.64} & 100.0& \textbf{23} & 0.36 & 0.21 & 0.65 &  & 64.41 & 89.7 & 100.0& 17 & 0.42 & 0.21 & 2.27 \\
\cdashline{1-17}
\rc\sigmazero & \multirow{-8}{*}{\textit \jiangloneshort} & 49.74 & 73.75 & 100.0& 25 & 0.04 & 0.2 & 0.89 & \multirow{-8}{*}{\textit \xushort} & \textbf{65.96} & \textbf{90.95} & 100.0& \textbf{16} & 0.09 & 0.2 & 2.52 \\ \hline

{SF} &  & 11.19 & 11.19 & 56.56 & 3072 & 1.42 & 0.37 & 1.56 &  & 24.28 & 26.54 & 51.90 & 3072 & 0.58 & 0.33 & 0.52 \\
EAD &  & 10.42 & 19.09 & 100.0 & 146 & 0.26 & 0.77 & 1.58&  & 18.82 & 26.17 & 100.0& 144 & 0.11 & 0.79 & 0.52 \\
{PDPGD} &  & 5.23 & 5.23 & 100.0 & 3057 & 0.05 & 0.20 & 1.43&  & 14.29 & 14.29 & 90.95 & 3057 & 0.03 & 0.2 & 0.47 \\
{VFGA} &  & 77.22 & 93.44 & 98.99 & 11 & 0.17 & 0.38 & 12.08&  & 48.49 & 74.14 & 94.16 & 26 & 0.12 & 0.73 & 3.18 \\
{FMN} &  & 89.83 & 97.72 & 98.86 & \textbf{8} & 0.05 & 0.20 & 1.43 &  & 46.75 & 69.77 & 80.68 & 27 & 0.03 & 0.2 & 0.48 \\
{BB} &  & 84.42 & 97.55 & 100.0 & 10 & 0.62 & 0.95 & 1.59 &  & \textbf{63.70} & 89.39 & 100.0& \textbf{17} & 0.43 & 1.13 & 0.53 \\
{BBadv} &  & 83.81 & 97.35 & 100.0 & 10 & 0.45 & 0.21 & 1.59 &  & 63.29 & \textbf{90.08} & 100.0& \textbf{17} & 0.35 & 0.21 & 0.53 \\
\cdashline{1-17}
\rc\sigmazero & \multirow{-8}{*}{\textit \standardshort} & \textbf{91.54} & \textbf{99.84} & 100.0 & 9 & 0.08 & 0.20 & 1.83 & \multirow{-8}{*}{\textit \addepallishort} & 60.79 & 86.02 & 100.0& 18 & 0.04 & 0.2 & 0.77 \\
\bottomrule
\end{tabular}
}
\end{table*}

\begin{table*}[htb]
\caption{Minimum-norm comparison results for ImageNet with $N=100$. See the caption of \autoref{tab:1k_remaining_minimum_norm_cifar_imagenet} for further details.}
\label{tab:0.1k_minimum_norm_imagenet}
\centering
\resizebox{1\textwidth}{!}{%
\renewcommand{\arraystretch}{1.1}
  \setlength\tabcolsep{1.5pt} 
\begin{tabular}{r|c|cccc|ccc|c|cccc|ccc}
\toprule
\multicolumn{1}{c}{\textbf{Attack}} & \multicolumn{1}{c}{\textbf{M}}& \multicolumn{1}{c}{\textbf{ASR$_{24}$}} & \multicolumn{1}{c}{\textbf{ASR$_{50}$}} & \multicolumn{1}{c}{\textbf{ASR$_\infty$}} & \multicolumn{1}{c}{\medianlzero} & \multicolumn{1}{c}{$\mathbf{s}$} & \multicolumn{1}{c}{$\mathbf{q}$}  & \multicolumn{1}{c}{\textbf{VRAM}} & \multicolumn{1}{c}{\textbf{M}} & \multicolumn{1}{c}{\textbf{ASR$_{24}$}} & \multicolumn{1}{c}{\textbf{ASR$_{50}$}} & \multicolumn{1}{c}{\textbf{ASR$_\infty$}} & \multicolumn{1}{c}{\medianlzero} & \multicolumn{1}{c}{$\mathbf{s}$} & \multicolumn{1}{c}{$\mathbf{q}$}  & \multicolumn{1}{c}{\textbf{VRAM}}\\
\hline\hline
\multicolumn{17}{c}{ImageNet} \\ \hline\hline
{EAD} &  & 34.7 & 35.9 & 100.0 & 484 & 1.02 & 0.67 & 0.46 &  & 32.4 & 33.0 & 100.0& 808 & 5.15 & 0.7 & 1.68 \\%
{VFGA} &  & 58.3 & 72.2 & 85.3 & 14 & 1.06 & 0.70 & \minfty &  & 40.0 & 46.8 & 56.9 & 66.5 & 9.23 & 1.2 & \minfty \\%
{FMN} &  & 55.4 & 64.5 & 68.1 & 14 & 0.08 & 0.20 & 0.66 &  & 39,9 & 46.2 & 47.5 & \binfty & 0.44 & 0.2 & 2.97 \\%
{BBadv} &  & 67.6 & 83.3 & 100.0 & 10 & 23.02 & 0.21 & 0.72 &  & \textbf{46.4} & \textbf{58.0} & 99.9 & 32 & 21.23 & 0.21 & 3.07 \\%
\cdashline{1-17}
\rc\sigmazero & \multirow{-5}{*}{\textit \resnetvulnerableshort} & \textbf{69.2} & \textbf{86.9} & 100.0 & \textbf{10} & 0.13 & 0.20 & 0.84 & \multirow{-5}{*}{\textit \debenedettishort} & 43.7 & 55.2 & 100.0& \textbf{32} & 0.61 & 0.2 & 3.20  \\ \hline%

{EAD} &  & 47.1 & 50.1 & 100.0 & 48 & 2.32 & 0.68 & 1.42 &  & 56.2 & 60.2 & 100.0 & 0 & 2.46 & 0.72 & 1.41 \\
{VFGA} &  & 54.3 & 63.2 & 96.7 & 13 & 2.88 & 0.72 & \minfty &  & 68.9 & 76.0 & 83.0 & 0 & 2.33 & 0.59 & \minfty \\
{FMN} &  & 55.9 & 60.0 & 62.4 & 10 & 0.20 & 0.20 & 2.30 &  & 67.8 & 72.0 & 74.3 & 0 & 0.20 & 0.20 & 2.30 \\
{BBadv} &  & 70.1 & 80.1 & 100.0 & 5 & 20.49 & 0.21 & 2.40 &  & 80.8 & 87.8 & 100.0 & 0 & 18.60 & 0.21 & 2.41 \\
\cdashline{1-17}
\rc\sigmazero & \multirow{-5}{*}{\textit \engstromimagenetshort} & \textbf{71.0} & \textbf{82.7} & 100.0 & \textbf{4} & 0.29 & 0.20 & 2.52 & \multirow{-5}{*}{\textit \wongshort} & \textbf{81.8} & \textbf{89.3} & 100.0 & 0 & 0.29 & 0.20 & 2.52 \\ \hline

{EAD} &  & 26.9 & 27.7 & 100.0& 1108 & 0.58 & 0.61 & 1.41 &  & 57.4 & 60.0 & 100.0 & 0 & 1.03 & 0.72 & 0.48 \\
{VFGA}  &  & 47.0 & 58.7 & 74.0 & 31 & 3.07 & 0.96 & \minfty &  & 66.8 & 75.2 & 83.9 & 0 & 0.91 & 0.59 & \minfty \\
{FMN}  &  & 44.4 & 50.6 & 53.2 & 47 & 0.16 & 0.2 & 2.30&  & 69.2 & 74.9 & 77.2 & 0 & 0.07 & 0.2 & 0.67 \\
{BBadv}  &  & \textbf{53.6} & \textbf{74.7} & 100.0& \textbf{20} & 23.86 & 0.21 & 2.41&  & 80.1 & 89.1 & 100.0& 0 & 19.68 & 0.21 & 0.73 \\
\cdashline{1-17}
\rc\sigmazero & \multirow{-5}{*}{\textit \hendrycksshort} & 52.9 & 74.4 & 100.0& 21 & 0.23 & 0.2 & 2.52  & \multirow{-5}{*}{\textit \salmanshort} & \textbf{82.2} & \textbf{90.5} & 100.0& 0 & 0.12 & 0.2 & 0.84 \\ 
\bottomrule
\end{tabular}
}
\end{table*}


\begin{table}[htb]
\caption{Fixed-budget comparison results with $N=1000$ on CIFAR10 remaining models. Sparse-RS was executed with double the steps, $2N$, to ensure fair comparison as it lacks backward passes. For each attack, we report the corresponding ASR with different feature budget levels (24,50,100). \rebuttal{We report the execution time  $\textbf{s}_{24}$ and query usage $\textbf{q}_{24}$ for the smaller $k=24$, as it requires, on average, more iterations due to the more challenging problem. Lastly we indicate with \sigmazero$^*$ the case where we use $\sigma = 1$ and $\tau_0 = 0.1$.}}
\label{tab:1k_remaining_maximum_budget_cifar}
\renewcommand{\arraystretch}{1.1}

  \centering
\resizebox{1.0\textwidth}{!}{%
  \setlength\tabcolsep{1.5pt} 
\setlength{\dashlinedash}{5pt}
\setlength{\dashlinegap}{3pt}
\begin{tabular}{c|c|ccc|ccc|c|ccc|ccc}
\toprule
\multicolumn{1}{c}{\textbf{Attack}} & \multicolumn{1}{c}{\textbf{M}} & \multicolumn{1}{c}{\textbf{ASR$_{24}$}} & \multicolumn{1}{c}{\textbf{ASR$_{50}$}} & \multicolumn{1}{c}{\textbf{ASR$_{100}$}} & \multicolumn{1}{c}{\textbf{q$_{24}$}} & \multicolumn{1}{c}{\textbf{s$_{24}$}} & \multicolumn{1}{c}{\textbf{VRAM}} & \multicolumn{1}{c}{\textbf{M}} & \multicolumn{1}{c}{\textbf{ASR$_{24}$}} & \multicolumn{1}{c}{\textbf{ASR$_{50}$}} & \multicolumn{1}{c}{\textbf{ASR$_{100}$}} & \multicolumn{1}{c}{\textbf{q$_{24}$}} & \multicolumn{1}{c}{\textbf{s$_{24}$}} & \multicolumn{1}{c}{\textbf{VRAM}} \\

\hline\hline
\multicolumn{15}{c}{CIFAR-10} \\ \hline\hline
\PGDlzero & \multirow{5}{*}{\standardshort} & 68.60 & 88.89 & 98.14 & 2.00 & 1.95 & 1.89 & \multirow{5}{*}{\chenshort} & 42.81 & 66.19 & 90.49 & 2.00 & 3.24 & 7.36 \\
Sparse-RS &  & 99.71 & 100.0 & 100.0 & 0.08 & 0.10 & 1.91 & & 72.54 & 86.72 & 94.84 & 0.78 & 1.10 & 7.35 \\
\SPGDproj &  & \textbf{99.82} & 100.0 & 100.0 & 0.02 & 0.16 & 2.06  & & 68.47 & 90.47 & 99.56 & 0.70 & 1.48 & 7.62 \\
\SPGDunproj &  & 97.84 & 99.98 & 100.0 & 0.09 & 0.37 & 2.06 & & 73.55 & 94.55 & 99.97 & 0.60 & 1.65 & 7.62\\
\cdashline{1-15} \cdashline{1-15} \rc\sigmazero & & 99.20 & 100.0 & 100.0 & 0.22 & 0.11 & 2.07 & & \textbf{81.23} & \textbf{97.33} & \textbf{99.97} & 0.52 & 0.54 & 7.76 \\ \hline

\PGDlzero & \multirow{5}{*}{\gowalshort} & 32.80 & 50.53 & 77.06 & 2.00 & 4.88 & 12.79 & \multirow{5}{*}{\xushort} & 31.45 & 52.79 & 80.27 & 2.00  & 2.01 & 2.91\\
Sparse-RS &  & \textbf{76.61} & 89.88 & 96.22 & 0.67 & 1.98 & 12.74 & & 68.77 & 82.06 & 89.81 & 0.85 & 0.56 & 2.89 \\
\SPGDproj &  & 63.66 & 87.07 & 98.67 & 0.80 & 2.83 & 13.77 & & 61.0 & 83.49 & 96.76 & 0.87 & 0.88 & 3.03 \\
\SPGDunproj &  & 64.28 & 88.25 & 99.09 & 0.77 & 2.77 & 13.77  & & 63.48 & 87.59 & 98.49 & 0.81 & 0.85 & 3.03 \\
\cdashline{1-15} \cdashline{1-15} \rc\sigmazero &  & 75.63 & \textbf{94.47} & \textbf{99.78} & 0.66 & 1.75 & 13.82 & & \textbf{79.59} & \textbf{96.93} & \textbf{99.91} & 0.57 & 2.04 & 2.91 \\ \hline

\PGDlzero & \multirow{5}{*}{\engstromshort} & 37.91 & 68.90 & 95.31 & 2.00 & 1.96 & 2.47 & \multirow{5}{*}{\addepallishort} & 38.33 & 61.88 & 89.50 & 2.00 & 1.12 & 0.51 \\
Sparse-RS &  & 63.75 & 84.49 & 95.74 & 0.97 & 0.61 & 2.46 &  & 64.80 & 81.46 & 91.13 & 0.91 & 0.45 & 0.50 \\
\SPGDproj &  & 72.82 & 96.74 & 99.98 & 0.61 & 0.94 & 2.57 & & 59.94 & 84.87 & 98.82 & 0.87 & 0.44 & 0.55 \\
\SPGDunproj &  & \textbf{81.64} & 99.09 & 100.0 & 0.42 & 0.69 & 2.57 & & 65.07 & 90.78 & \textbf{99.83} & 0.75 & 0.41 & 0.55 \\
\cdashline{1-15} \rc\sigmazero &  & 81.38 & \textbf{99.15} & 100.0 & 0.46 & 0.21 & 2.68 & & \textbf{73.96} & \textbf{94.21} & 99.80 & 0.67 & 0.14 & 0.57 \\ \hline
Sparse-RS & \multirow{5}{*}{\zhongtradesshort} & 28.08 & 41.89 & 58.45 & 0.38 & 1.53 & 0.59 & \multirow{5}{*}{\zhongsatshort} & 51.64 & 71.27 & 86.57 & 0.32 & 1.14 & 0.59 \\
\SPGDunproj &  & 15.87 & 21.43 & 32.67 & 0.26 & 1.71 & 0.65 &  & 22.78 & 26.56 & 34.09 & 0.31 & 1.58 & 0.64 \\
\SPGDproj &  & 13.61 & 17.07 & 30.11 & 0.25 & 1.74 & 0.65 &  & 24.52 & 34.39 & 59.89 & 0.30 & 1.54 & 0.65 \\
\sigmazero &  & 12.38 & 15.91 & 30.43 & 0.20 & 1.77 & 1.03 &  & 18.52 & 21.31 & 28.81 & 0.27 & 1.65 & 1.03 \\
\cdashline{1-15}\rc\sigmazero$^*$ &  & \textbf{44.78} & \textbf{85.05} & \textbf{99.76} & 0.15 & 1.33 & 1.03 &  & \textbf{54.62} & \textbf{90.12} & \textbf{99.94} & 0.13 & 1.15 & 1.03 \\

\bottomrule
\end{tabular}}
\end{table}
\begin{table*}[htb]
\centering
\caption{\rebuttal{Fixed-budget comparison results for ImageNet with $N=1000$ on remaining models. Sparse-RS was executed with double the steps, $2N$, to ensure fair comparison as it lacks backward passes. For each attack, we report the corresponding ASR with budget level $k=150$. We report the execution time  $\textbf{s}_{100}$ and query usage $\textbf{q}_{100}$ for the smaller $k=100$, as it requires, on average, more iterations due to the more challenging problem.}
}
\label{tab:1k_remaining_maximum_budget_imagenet}
\centering
  \setlength\tabcolsep{2.7pt} 
\setlength{\dashlinedash}{5pt}
\setlength{\dashlinegap}{3pt}
\begin{tabular}{c|c|cc|ccc|c|cc|ccc}
\toprule
\multicolumn{1}{c}{\textbf{Attack}} & \multicolumn{1}{c}{\textbf{M}} & \multicolumn{1}{c}{\textbf{ASR$_{100}$}} & \multicolumn{1}{c}{\textbf{ASR$_{150}$}} & \multicolumn{1}{c}{\textbf{q$_{100}$}} & \multicolumn{1}{c}{\textbf{s$_{100}$}} & \multicolumn{1}{c}{\textbf{VRAM}} & \multicolumn{1}{c}{\textbf{M}} & \multicolumn{1}{c}{\textbf{ASR$_{150}$}} & \multicolumn{1}{c}{\textbf{ASR$_{150}$}} & \multicolumn{1}{c}{\textbf{q$_{100}$}} & \multicolumn{1}{c}{\textbf{s$_{100}$}} & \multicolumn{1}{c}{\textbf{VRAM}} \\
\hline\hline
\multicolumn{13}{c}{ImageNet} \\ \hline\hline

Sparse-RS & \multirow{4}{*}{\wongshort} & 83.6 & 87.5 & 0.44 & 2.85 & 4.39 & \multirow{4}{*}{\salmanshort} & 85.4 & 89.2 & 0.41 &  3.46 &  1.29 \\
\SPGDproj &  & 89.8 & 94.5 & 0.24 & 1.64 &  4.48 &  & 90.4 & 95.2 & 0.22 & 0.98 & 1.33 \\
\SPGDunproj &  & 86.5 & 92.6 & 0.29 & 1.55 &  4.48 &  & 89.1 & 94.0 & 0.24 &  1.15 &  1.33\\
\cdashline{1-13} \rc\sigmazero & & \textbf{95.9} & \textbf{98.2} & 0.12 & 0.16 & 4.90 & & \textbf{98.1} & \textbf{98.8} & 0.10 & 0.08 & 1.79 \\ \hline

    Sparse-RS & \multirow{4}{*}{\pengshort} & 58.20 & 60.60 & 0.95 & 5.21 & 17.43 & \multirow{4}{*}{\moshort} & 49.20 & 52.10 & 1.13 & 3.28 & 7.89 \\
    \SPGDproj &  & 67.50 & 75.50 & 0.70 & 4.85 & 17.80 &  & 65.10 & 75.20 & 0.75 & 3.56 & 8.24 \\
    \SPGDunproj &  & 65.70 & 75.10 & 0.73 & 5.37 & 17.82 &  & 65.10 & 75.20 & 0.73 & 5.68 & 8.23 \\
    \cdashline{1-13}\rc\sigmazero &  & \textbf{82.10} & \textbf{87.00} & 0.43 & 1.87 & 19.03 &  & \textbf{78.00} & \textbf{86.20} & 0.50 & 1.67 & 9.46 \\
    
\bottomrule
\end{tabular}
\end{table*}

\begin{table}[htb]
\caption{Fixed-budget comparison results with $N=5000$ on MNIST. See the caption of Table~\ref{tab:1k_remaining_maximum_budget_cifar} for further details.}
\label{tab:5k_maximum_budget_mnist}
\renewcommand{\arraystretch}{1.1}
\centering
\resizebox{1.0\textwidth}{!}{%
  \setlength\tabcolsep{1.5pt} 
\setlength{\dashlinedash}{5pt}
\setlength{\dashlinegap}{3pt}
  
\begin{tabular}{c|c|ccc|ccc|c|ccc|ccc}
\toprule
\multicolumn{1}{c}{\textbf{Attack}} & \multicolumn{1}{c}{\textbf{M}} & \multicolumn{1}{c}{\textbf{ASR$_{24}$}} & \multicolumn{1}{c}{\textbf{ASR$_{50}$}} & \multicolumn{1}{c}{\textbf{ASR$_{100}$}} & \multicolumn{1}{c}{\textbf{q$_{24}$}} & \multicolumn{1}{c}{\textbf{s$_{24}$}} & \multicolumn{1}{c}{\textbf{VRAM}} & \multicolumn{1}{c}{\textbf{M}} & \multicolumn{1}{c}{\textbf{ASR$_{24}$}} & \multicolumn{1}{c}{\textbf{ASR$_{50}$}} & \multicolumn{1}{c}{\textbf{ASR$_{100}$}}& \multicolumn{1}{c}{\textbf{q$_{24}$}} & \multicolumn{1}{c}{\textbf{s$_{24}$}} & \multicolumn{1}{c}{\textbf{VRAM}}  \\ \hline\hline
\multicolumn{15}{c}{MNIST} \\ \hline\hline

\hline
Sparse-RS & \multirow{4}{*}{\smallcnnddnshort} & 88.13 & 99.26 & 99.99 & 2.45 & 1.86 & 0.04 & \multirow{4}{*}{\smallcnntradesshort} & \textbf{99.88} & 99.97 & 100.0 & 0.31 & 0.17 & 0.04 \\
\SPGDproj &  & 81.22 & 99.30 & 100.0 & 2.83 & 1.50 & 0.05 & & 83.88 & 99.88 & 99.97 & 2.33 & 0.9 & 0.05 \\
\SPGDunproj &  & 87.30 & 99.85 & 100.0 & 1.60 & 1.44 & 0.05 & & 74.38 & 99.46 & 99.99 & 3.47 & 0.96 & 0.05\\
\cdashline{1-15} \rc\sigmazero &  & \textbf{88.63} & \textbf{100.0} & 100.0 & 1.38 & 0.20 & 0.08 & & 99.67 & \textbf{100.0} & 100.0 & 0.24 &0.02 & 0.08 \\ \hline
\end{tabular}}
\end{table}
\begin{table}[htbp]
\caption{Fixed-budget comparison results with $N=5000$ on CIFAR10. See the caption of Table~\ref{tab:1k_remaining_maximum_budget_cifar} for further details.}
\label{tab:5k_maximum_budget_cifar}
\renewcommand{\arraystretch}{1.1}
\centering
\resizebox{1.0\textwidth}{!}{%
  \setlength\tabcolsep{1.5pt} 
\setlength{\dashlinedash}{5pt}
\setlength{\dashlinegap}{3pt}
\centering
\begin{tabular}{c|c|ccc|ccc|c|ccc|ccc}
\toprule
\multicolumn{1}{c}{\textbf{Attack}} & \multicolumn{1}{c}{\textbf{M}} & \multicolumn{1}{c}{\textbf{ASR$_{24}$}} & \multicolumn{1}{c}{\textbf{ASR$_{50}$}} & \multicolumn{1}{c}{\textbf{ASR$_{100}$}} & \multicolumn{1}{c}{\textbf{q$_{24}$}} & \multicolumn{1}{c}{\textbf{s$_{24}$}} & \multicolumn{1}{c}{\textbf{VRAM}} & \multicolumn{1}{c}{\textbf{M}} & \multicolumn{1}{c}{\textbf{ASR$_{24}$}} & \multicolumn{1}{c}{\textbf{ASR$_{50}$}} & \multicolumn{1}{c}{\textbf{ASR$_{100}$}}  & \multicolumn{1}{c}{\textbf{q$_{24}$}} & \multicolumn{1}{c}{\textbf{s$_{24}$}} & \multicolumn{1}{c}{\textbf{VRAM}} \\

\hline\hline
\multicolumn{15}{c}{CIFAR-10} \\ \hline\hline

Sparse-RS & \multirow{4}{*}{\carmonshort} & \textbf{82.94} & 94.77 & 98.68 & 2.55 & 1.81 & 1.92 & \multirow{4}{*}{\standardshort} & \textbf{99.93} & 99.98 & 99.99 & 0.15 & 0.14 & 1.91  \\%
\SPGDproj &  & 71.88 & 93.17 & 99.75 & 3.21 & 2.78 & 2.06 &  & 99.93 & 100.0 & 100.0 & 0.17 & 0.18 & 2.05 \\%
\SPGDunproj &  & 69.73 & 92.86 & 99.79 & 3.34 & 2.98 & 2.06 & & 99.73 & 100.0 & 100.0 & 0.26 & 1.31 & 2.05 \\%
\cdashline{1-15} \rc\sigmazero &  & 80.91 & \textbf{96.81} & \textbf{99.98} & 3.23 & 1.35 & 2.09 & & 99.81 & 100.0 & 100.0 & 0.61 & 0.18 & 2.05  \\ \hline%

Sparse-RS & \multirow{4}{*}{\augustinshort} & 71.21 & 90.28 & 97.69 & 3.89 & 2.49 & 2.46 & \multirow{4}{*}{\engstromshort} & 77.06 & 94.33 & 99.25 & 3.38 & 2.15 & 2.46 \\%
\SPGDproj &  & 64.88 & 92.03 & 99.85 & 3.97 & 3.69 & 2.57 & & 78.41 & 98.36 & 100.0 & 2.59 & 5.87 & 2.57 \\%
\SPGDunproj &  & 68.61 & 94.99 & 99.96 & 3.49 & 3.46 & 2.57 & & \textbf{84.17} & 99.41 & 100.0 & 1.85 & 6.26 & 2.57  \\
\cdashline{1-15} \rc\sigmazero &  & \textbf{78.14} & \textbf{98.39} & \textbf{100.0} & 3.16 & 1.02 & 2.70 & & 83.99 & \textbf{99.49} & 100.0 & 1.82 & 0.69 & 2.70  \\  \hline

Sparse-RS & \multirow{4}{*}{\croceloneshort} & 38.43 & 58.27 & 79.59 & 6.77 & 2.61 & 0.69 & \multirow{4}{*}{\xushort} & 79.69 & 89.98 & 95.3 & 2.81 & 1.95 & 2.89\\
\SPGDproj &  & 34.62 & 65.54 & 96.55 & 6.70 & 2.52 & 0.73 & & 66.39 & 87.65 & 98.19 & 3.73 & 2.75 & 3.03 \\
\SPGDunproj  &  & 37.29 & 72.03 & 98.49 & 6.48 & 2.96 & 0.73 & & 68.03 & 90.7 & 99.23 & 3.56 & 2.64 & 3.04 \\
\cdashline{1-15} \rc\sigmazero &  & \textbf{40.99} & \textbf{76.00} & \textbf{98.98} & 6.32 & 1.42 & 0.77 & & \textbf{83.92} & \textbf{98.39} & \textbf{99.99} & 2.19 & 0.99 & 3.09 \\ \hline

Sparse-RS & \multirow{4}{*}{\jiangloneshort} & 54.85 & 71.95 & 86.25 & 5.03  & 2.37  & 0.69  & \multirow{4}{*}{\addepallishort} & 76.62 & 91.5 & 97.89 & 3.20  & 1.46  & 0.50 \\
\SPGDproj &  & 53.36 & 80.97 & 99.13 & 4.94 & 2.24  & 0.73 & & 65.92 & 90.38 & 99.72 & 3.86  & 1.63  & 0.55 \\
\SPGDunproj &  & \textbf{57.31} & \textbf{86.11} & \textbf{99.72} & 4.46 & 2.36  & 0.73  & & 68.59 & 92.93 & 99.91 & 3.71  & 1.62  & 0.55 \\
\cdashline{1-15} \rc\sigmazero &  & 57.11 & 84.54 & 99.34 & 4.47 & 1.04  & 0.77  & & \textbf{77.74} & \textbf{95.86} & \textbf{99.92} & 2.75  & 0.57  & 0.59 \\
\bottomrule
\end{tabular}}

\end{table}
\begin{table}[htbp]
\caption{Fixed-budget comparison results with $N=5000$ on ImageNet. See the caption of \autoref{tab:1k_remaining_maximum_budget_imagenet} for further details.}
\label{tab:5k_maximum_budget_imagenet}
  \setlength\tabcolsep{9pt} 
  \centering

\renewcommand{\arraystretch}{1.1}

\centering
\resizebox{1.0\textwidth}{!}{%
  \setlength\tabcolsep{1.5pt} 
\setlength{\dashlinedash}{5pt}
\setlength{\dashlinegap}{3pt}

    \begin{tabular}{c|c|cc|ccc|c|cc|ccc}
    \toprule
    \multicolumn{1}{c}{\textbf{Attack}} & \multicolumn{1}{c}{\textbf{M}} & \multicolumn{1}{c}{\textbf{ASR$_{100}$}} & \multicolumn{1}{c}{\textbf{ASR$_{150}$}} & \multicolumn{1}{c}{\textbf{q$_{100}$}} &\multicolumn{1}{c}{\textbf{s$_{100}$}} &\multicolumn{1}{c}{\textbf{VRAM}} & \multicolumn{1}{c}{\textbf{M}} & \multicolumn{1}{c}{\textbf{ASR$_{100}$}} & \multicolumn{1}{c}{\textbf{ASR$_{150}$}} &\multicolumn{1}{c}{\textbf{q$_{100}$}} &\multicolumn{1}{c}{\textbf{s$_{100}$}} &\multicolumn{1}{c}{\textbf{VRAM}} \\ 
    \hline\hline
\multicolumn{13}{c}{ImageNet} \\ \hline\hline
    
    Sparse-RS & \multirow{4}{*}{\resnetvulnerableshort} & 94.2 & 95.1 & 1.39 & 7.73 & 1.29 & \multirow{4}{*}{\debenedettishort} & 48.8 & 51.7 & 5.68 & 13.51 & 5.73 \\%
    \SPGDproj & & 97.3 & 99.6 & 0.45 & 1.95 & 1.41 & & 66.2 & 78.4 & 3.68 &21.65 & 5.84 \\%
    \SPGDunproj &  & 93.6 & 98.5 & 0.71 & 2.48 & 1.40 & & 64.1 & 78.5 & 3.78 & 20.71 & 5.84\\%
    \cdashline{1-9} \rc\sigmazero & & \textbf{100.0} & \textbf{100.0} & 0.72 & 0.31 & 1.83 & & \textbf{77.3} & \textbf{87.8} & 2.42 & 4.17 & 6.33 \\ \hline

    Sparse-RS & \multirow{4}{*}{\engstromimagenetshort} & 85.1 & 86.8 & 2.06 & 11.0 & 4.39 & \multirow{4}{*}{\wongshort} & 89.6 & 91.3 & 1.54 & 3.35 & 4.39\\
    \SPGDproj &  & 85.9 & 92.8 & 1.63 & 8.33 & 4.49 & & 92.0 & 95.9 & 0.94 & 3.35 & 4.39\\
    \SPGDunproj &  & 81.3 & 90.5 & 2.00 & 6.88 & 4.49 & & 88.0 & 93.1 & 1.28 & 5.89 & 4.48 \\
    \cdashline{1-9} \rc\sigmazero &  & \textbf{94.6} & \textbf{97.3} & 0.63 & 0.61 & 4.94 & & \textbf{96.9} & \textbf{98.4} & 0.41 & 0.34 & 4.94 \\ \hline

    Sparse-RS & \multirow{4}{*}{\hendrycksshort} & 74.8 & 76.6 & 3.54 & 6.38 & 4.39 & \multirow{4}{*}{\salmanshort} & 87.5 & 92.7 & 1.36 &2.05 & 1.29 \\
    \SPGDproj &  & 87.6 & 95.4 & 1.61 & 6.29 & 4.49 & & 96.5 & 97.0 & 0.83 & 1.39 & 1.33 \\
    \SPGDunproj & & 81.4 & 93.7 & 2.04 & 6.54 & 4.49 & & 90.4 & 94.7 & 2.04 &2.35 & 1.33 \\
    \cdashline{1-9} \rc\sigmazero & & \textbf{98.2} & \textbf{99.7} & 1.68  & 1.44 & 4.94 & & \textbf{97.2} & \textbf{99.1} & 0.35 & 0.12 & 1.83 \\

    \bottomrule
    \end{tabular}}

\end{table}

\begin{table}[htbp]
\caption{Fixed-budget comparison results with $N=10000$ on CIFAR-10 and ImageNet. See the caption of Tables~ \ref{tab:1k_remaining_maximum_budget_cifar}-\ref{tab:1k_remaining_maximum_budget_imagenet} for further details.}
\label{tab:10k_steps_cifar_imgnet}
  \setlength\tabcolsep{9pt} 
  \centering
\renewcommand{\arraystretch}{1.1}

\centering
  \setlength\tabcolsep{1.2pt} 
\setlength{\dashlinedash}{5pt}
\setlength{\dashlinegap}{3pt}

\begin{tabular}{c|c|ccc|ccc|c|cc|ccc}
\toprule

\multicolumn{1}{c}{\textbf{Attack}} & \multicolumn{1}{c}{\textbf{M}} & \multicolumn{1}{c}{\textbf{ASR$_{24}$}} & \multicolumn{1}{c}{\textbf{ASR$_{50}$}} & \multicolumn{1}{c}{\textbf{ASR$_{100}$}} & \multicolumn{1}{c}{\textbf{q$_{24}$}} & \multicolumn{1}{c}{\textbf{s$_{24}$}} & \multicolumn{1}{c}{\textbf{VRAM}} & \multicolumn{1}{c}{\textbf{M}} & \multicolumn{1}{c}{\textbf{ASR$_{100}$}} & \multicolumn{1}{c}{\textbf{ASR$_{150}$}}  & \multicolumn{1}{c}{\textbf{q$_{100}$}} & \multicolumn{1}{c}{\textbf{s$_{100}$}} & \multicolumn{1}{c}{\textbf{VRAM}} \\ \hline

Sparse-RS & \multirow{4}{*}{\croceloneshort} & 41.12 & 63 & 83.99 & 3.44 & 12.82 & 0.60 & \multirow{4}{*}{\engstromimagenetshort} & 87.2 & 88.3 & 9.66 & 3.67 & 0.14 \\
\SPGDunproj &  & 37.95 & 72.6 & 98.59 & 1.99 & 12.56 & 0.65 &  & 86.9 & 91.6 & 7.84 & 1.95 & 0.15 \\
\SPGDproj &  & 35.89 & 67.92 & 97.42 & 2.12 & 13.2 & 0.65 &  & 81.8 & 88.9 & 8.4 & 3.85 & 0.15 \\
\cdashline{1-14}\rc\sigmazero &  & \textbf{41.67} & \textbf{76.38} & \textbf{99.01} & 0.33 & 3.07 & 2.41 &  & \textbf{95.1} & \textbf{97.2} & 1.31 & 0.23 & 0.21  \\
  \bottomrule
\end{tabular}

\end{table}

\clearpage

\section{Visual Comparison}\label{appendix:extra_images}
\begin{figure}[htbp]
    \centering
    \includegraphics[width=1\textwidth]{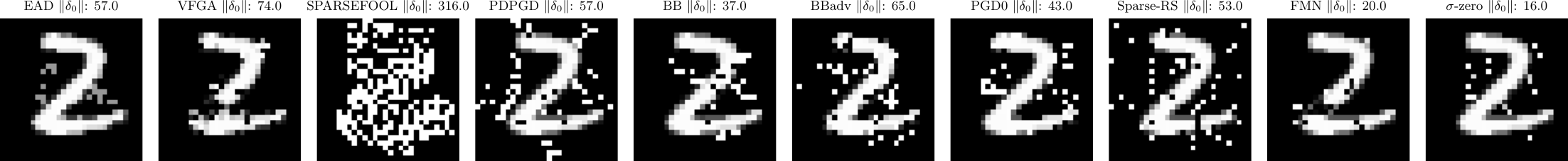}
    \includegraphics[width=1\textwidth]{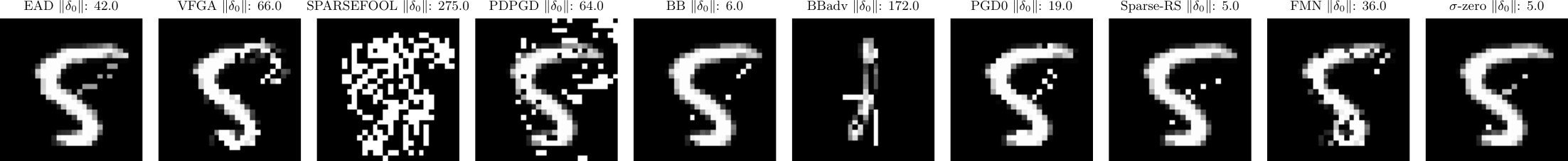}
    \includegraphics[width=1\textwidth]{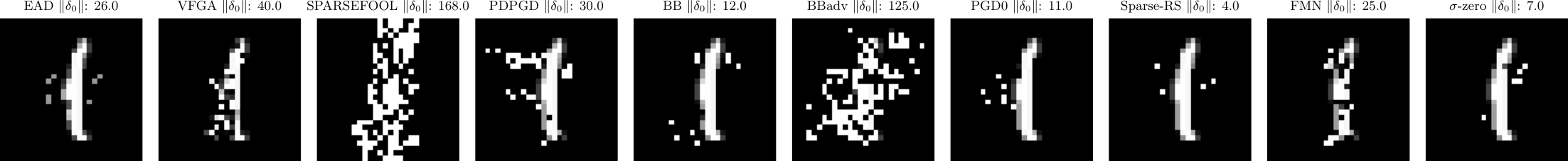}
    \includegraphics[width=1\textwidth]{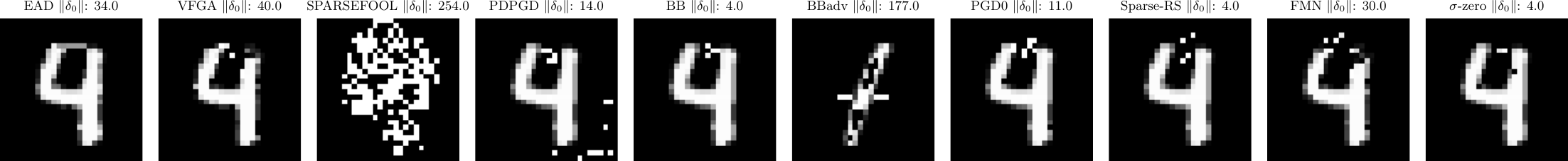}
        \caption{Randomly chosen adversarial examples from MNIST~\smallcnntradesshort.}
    \label{fig:adv_examples_mnist_appendix}
\end{figure}

\begin{figure}[htbp]
    \centering
    \includegraphics[width=1\textwidth]{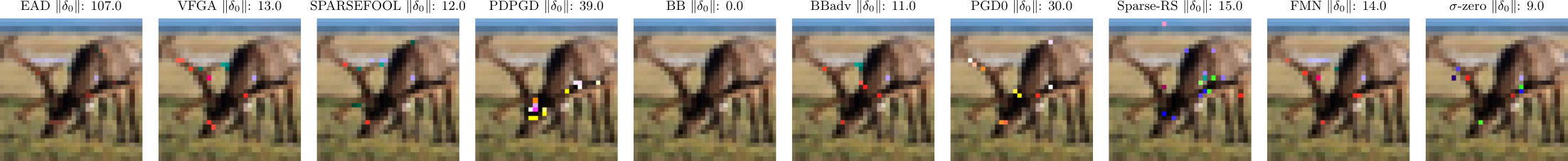}
    \includegraphics[width=1\textwidth]{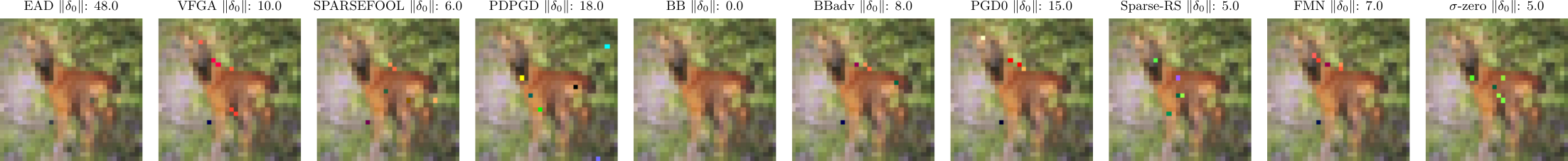}
    \includegraphics[width=1\textwidth]{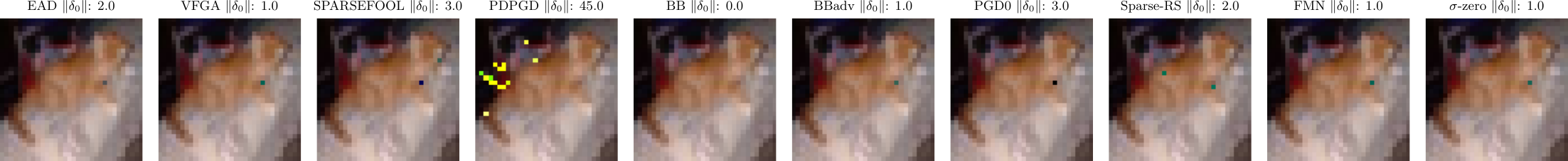}
    \includegraphics[width=1\textwidth]{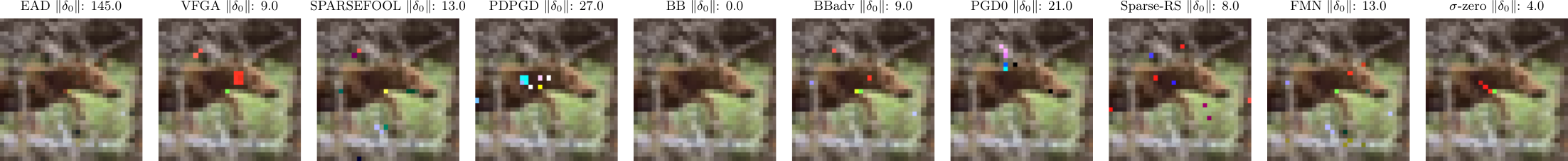}
    \caption{Randomly chosen adversarial examples from CIFAR-10~\carmonshort.}
    \label{fig:adv_examples_cifar_appendix}
\end{figure}

\begin{figure}[htbp]
    \centering
    \includegraphics[width=1\textwidth]{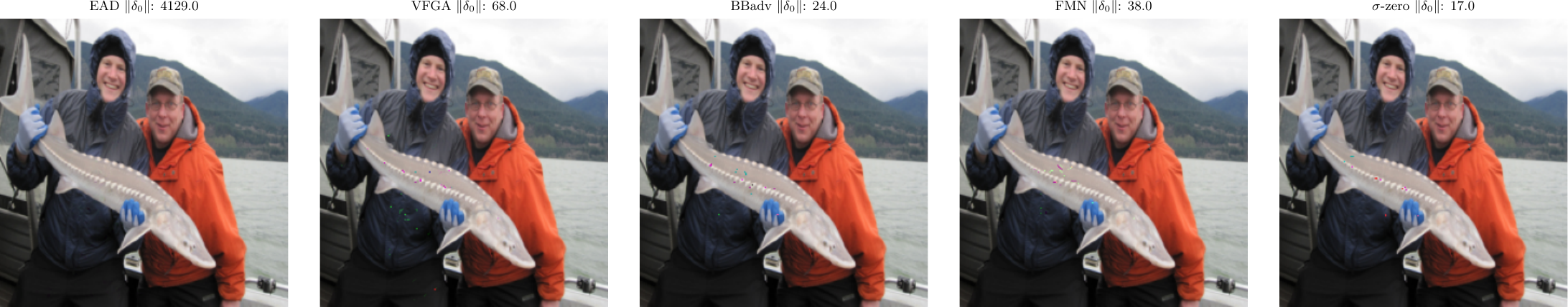}
    \includegraphics[width=1\textwidth]{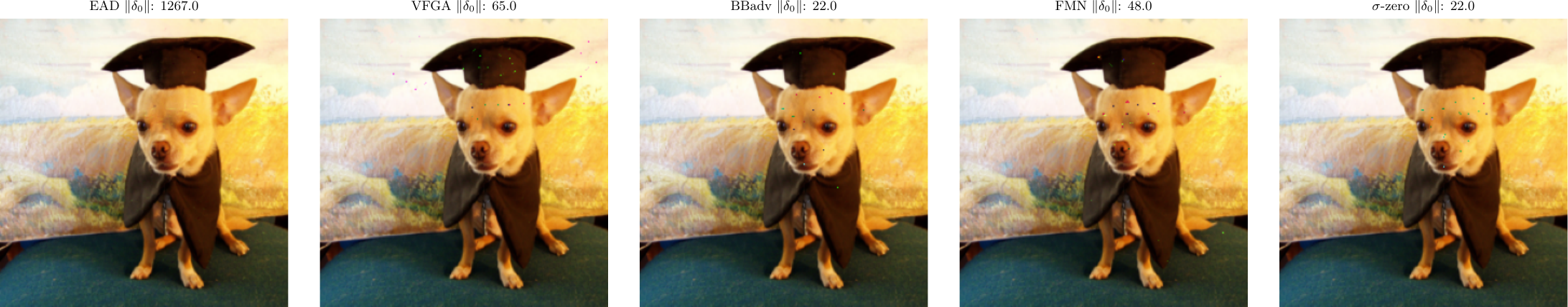}
    \includegraphics[width=1\textwidth]{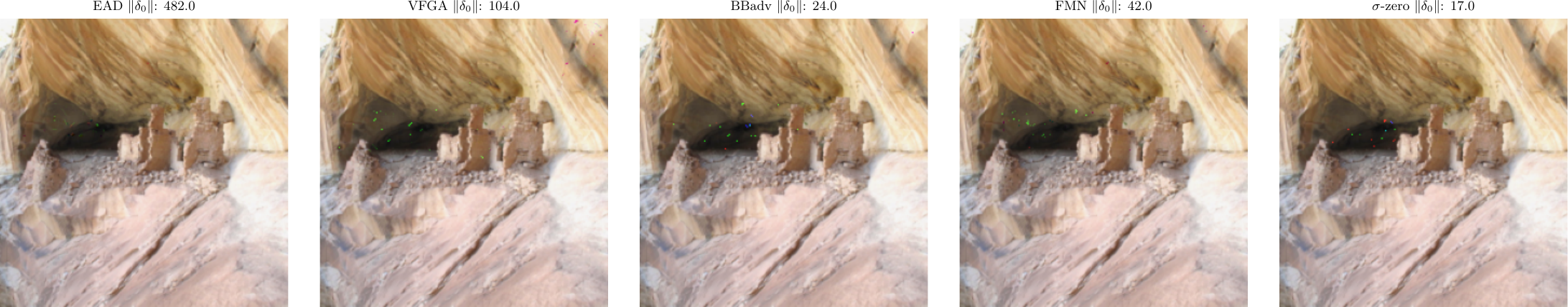}
    \includegraphics[width=1\textwidth]{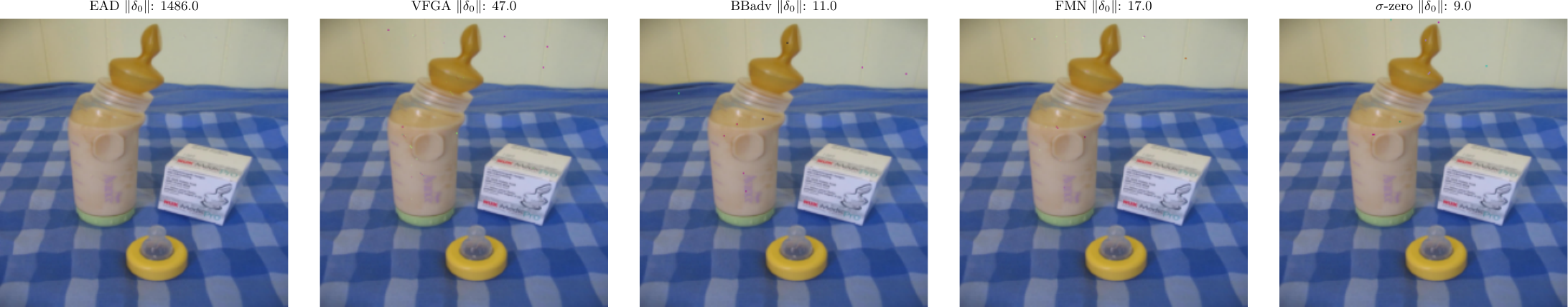}
    \includegraphics[width=1\textwidth]{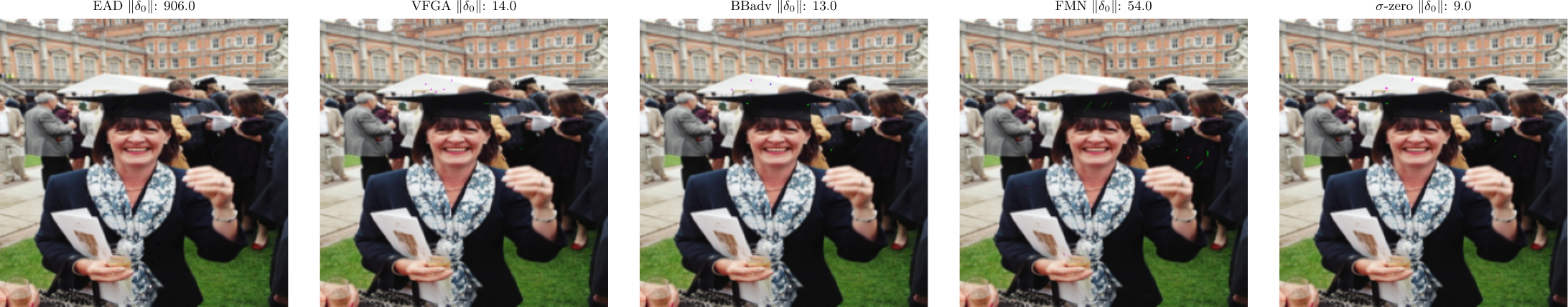}
    \includegraphics[width=1\textwidth]{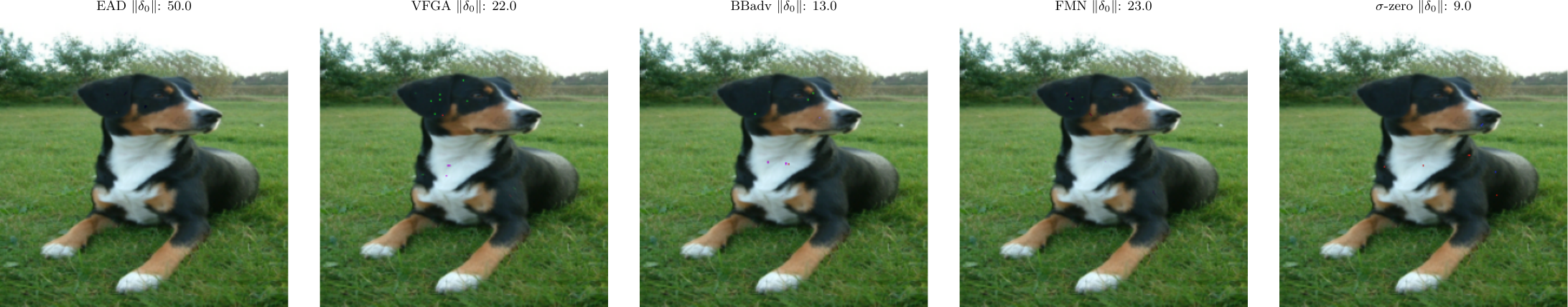}

    \caption{Randomly chosen adversarial examples from ImageNet~\resnetvulnerableshort.}
    \label{fig:adv_examples_imagenet_appendix}
\end{figure}

In Figures \ref{fig:adv_examples_mnist_appendix}-\ref{fig:adv_examples_imagenet_appendix}, we show adversarial examples generated with competing \lzero-attacks, and our \sigmazero. First, we can see that \lzero adversarial perturbations are clearly visually distinguishable~\cite{carlini-17-towards,Brendel2019BB,Pintor2021FMN}. Their goal, indeed, is not to be indistinguishable to the human eye -- a common misconception related to adversarial examples~\citep{Biggio18WildPatterns,gilmer18} -- but rather to show whether and to what extent models can be fooled by just changing a few input values.\\
A second observation derived from Figures \ref{fig:adv_examples_mnist_appendix}-\ref{fig:adv_examples_imagenet_appendix} is that the various attacks presented in the state of the art can identify distinct regions of vulnerability. For example, note how FMN and VFGA find similar perturbations, as they mostly target overlapping regions of interest. Conversely, EAD finds sparse perturbations scattered throughout the image but with a lower magnitude. This divergence is attributed to EAD's reliance on an \lone regularizer, which promotes sparsity, thus diminishing perturbation magnitude without necessarily reducing the number of perturbed features. Conversely, our attack does not focus on specific areas or patterns within the images but identifies diverse critical features, whose manipulation is sufficient to mislead the target models. Given the diverse solutions offered by the attacks, we argue that their combined usage may still improve adversarial robustness evaluation to sparse attacks.


\end{document}